%% file: main.tex
\newif\ifreview
\reviewtrue

\documentclass[10pt,conference,compsocconf,final]{IEEEtran}

\newcommand{\carml}{XSP\xspace}

\usepackage[bookmarks=false,bookmarksnumbered=true,
pdfpagemode={UseOutlines},plainpages=false,pdfpagelabels=true,
colorlinks=true,linkcolor={black},citecolor={black},urlcolor={black},breaklinks=true]{hyperref}

\IEEEoverridecommandlockouts

\DeclareUrlCommand{\bftturl}{}
\newcommand*\DeclareFancyUrlCommand[2]{%
    \expandafter\DeclareFancy@UrlCommand
    \expandafter{\csname fancyurl@\expandafter\@gobble\string#2\endcsname}{#1}{#2}%
}

\usepackage{enumitem}

\usepackage{adjustbox}
\usepackage{cite}
\usepackage{pstricks}
\usepackage{amsmath,amssymb,amsfonts}
\usepackage{algorithmic}
\usepackage{graphicx}
\usepackage{textcomp}
\usepackage{xcolor}

\usepackage{amsmath}

\usepackage[breakable, theorems, skins]{tcolorbox}
\tcbset{enhanced}

\usepackage{hyperref}

\usepackage{microtype}
\usepackage{graphicx}
\usepackage{verbatim}
\usepackage{booktabs} %
\usepackage{soul}
\usepackage{amsthm}
\usepackage{amsmath}
\usepackage{blindtext}
\usepackage{fancyhdr}
\usepackage{xspace}
\usepackage{multirow}
\usepackage{comment}
\usepackage{array}
\usepackage{makecell}
\usepackage{rotating}
\usepackage{filecontents}
\usepackage{pgfplots}
\usepackage{pgfplotstable}
\usepackage{tikz}
\usepackage[normalem]{ulem}
\usepackage[eulergreek]{sansmath}
\usepackage{listings}
\usepackage{xargs}                      %
\usepackage{amssymb}%
\usepackage{pifont}%
\usepackage{tcolorbox}
\usepackage{environ}
\usepackage{tabularx}%
\usepackage{arydshln}

\newcommand{\cmmnt}[1]{\ignorespaces}

\definecolor{myred}{rgb}{0.843137,0.188235,0.152941}
\definecolor{myblack}{rgb}{0.27451,0.32549,0.384314}
\definecolor{mygreen}{rgb}{0.301961,0.686275,0.290196}
\definecolor{myyellow}{rgb}{0.996078,0.878431,0.564706}
\definecolor{myblue}{rgb}{0.568627,0.74902,0.858824}

\pgfplotsset{compat=newest,}
\usepgfplotslibrary{external}
\usetikzlibrary{babel}
\usetikzlibrary{shapes, calc, shapes, arrows, snakes}
\usetikzlibrary{arrows}
\usetikzlibrary{shapes}

\definecolor{lightyellow}{RGB}{255, 250, 236}
\definecolor{textdark}{RGB}{100, 52, 20}
\definecolor{borderorange}{RGB}{253, 129, 36}
\definecolor{lightgray}{RGB}{214, 214, 214}
\definecolor{countrygray}{RGB}{153, 153, 153}

\pgfplotsset{every axis/.style={scale only axis}}

\pgfplotsset{%
    ,tick label style = {font=\small\sansmath\sffamily\footnotesize} %
    ,every axis label = {font=\small\sansmath\sffamily\footnotesize}
    ,legend style = {font=\tiny\sansmath\sffamily\tiny}
    ,label style = {font=\sansmath\sffamily\footnotesize}
}

\usepgfplotslibrary{colorbrewer}
\pgfplotsset{cycle list/Dark2-8}
\pgfplotsset{cycle list/RdYlBu-6} 
\pgfplotsset{cycle list/Set1}
\pgfplotsset{cycle list/Paired}

\newcommand{\ignore}[1]{}

\makeatletter
\newsavebox{\measure@tikzpicture}
\NewEnviron{scaletikzpicturetowidth}[1]{%
  \def\tikz@width{#1}%
  \begin{lrbox}{\measure@tikzpicture}%
  \BODY
  \end{lrbox}%
  \pgfmathparse{#1/\wd\measure@tikzpicture}%
  \BODY
}

\newcommand{\yes}{{\color{black} \ding{51}}}
\newcommand{\no}{{\color{black} \ding{55}}}

\newtcbox\circlebox{hbox, on line, colback=black, enhanced, frame hidden, boxrule=0pt, 
    top=-2pt, bottom=-2pt, right=-2pt, left=-2pt, rounded corners, arc=2pt}

\newtcbox\questionbox{hbox, on line, colback=black, enhanced, frame hidden, boxrule=0pt, 
    top=-2pt, bottom=-2pt, right=-2pt, left=-2pt, rounded corners, arc=2pt}

\DeclareRobustCommand\analysis[1]{\tikz[baseline=(char.base)]{
            \node[rounded corners,white,draw,fill=black,line width=1pt,rounded corners=2pt,inner sep=1.5pt,anchor=base] (char) {$\textsf{A}$\normalfont\sffamily\textsf{#1}};}\nolinebreak\ignorespacesafterend\hspace{-2pt}}

\newcommand*\circledwhite[1]{\tikz[baseline=(char.base)]{
            \node[shape=circle,draw,inner sep=0.0pt] (char) 
             {\small\sffamily\bfseries{#1}};}\nolinebreak\ignorespacesafterend\hspace{-2pt}}

\DeclareRobustCommand*\circled[1]{\tikz[baseline=(char.base)]{
            \node[shape=circle,white,draw,fill=black,inner sep=0.0pt] (char) 
             {\small\sffamily\bfseries{#1}};}\nolinebreak\ignorespacesafterend\hspace{-2pt}}
            
\def\BibTeX{{\rm B\kern-.05em{\sc i\kern-.025em b}\kern-.08em
    T\kern-.1667em\lower.7ex\hbox{E}\kern-.125emX}}

\let\oldbibliography\thebibliography
\renewcommand{\thebibliography}[1]{\oldbibliography{#1}
\setlength{\itemsep}{0pt}} %

\usepackage{setspace}

\usepackage[htt]{hyphenat}
\usepackage{xifthen}
\usepackage[normalem]{ulem}

\newcommand{\oneline}[1]{%
  \newdimen{\namewidth}%
  \setlength{\namewidth}{\widthof{#1}}%
  \ifthenelse{\lengthtest{\namewidth < \textwidth}}%
  {#1}%
  {\resizebox{\textwidth}{!}{#1}}%
}

\begin{document}

\title{XSP: Across-Stack Profiling and Analysis of Machine Learning Models on GPUs}

\author{
\IEEEauthorblockN{Cheng Li\textsuperscript{1*}, Abdul Dakkak\textsuperscript{1*}, Jinjun Xiong\textsuperscript{2}, Wei Wei\textsuperscript{3}, Lingjie Xu\textsuperscript{4}, Wen-mei Hwu\textsuperscript{1}}
\IEEEauthorblockA{\textsuperscript{1}University of Illinois Urbana-Champaign, \textsuperscript{2}IBM T. J. Watson Research Center, \textsuperscript{3}Alibaba, \textsuperscript{4}Biren Research (formerly at Alibaba) \\
\{cli99, dakkak, w-hwu\}@illinois.edu, jinjun@us.ibm.com, w.wei@alibaba-inc.com, lingjie@birentech.com}
}

\renewcommand{\thefootnote}{\fnsymbol{footnote}}

\maketitle

\footnotetext[1]{The two authors contributed equally to this paper.}

\input{sections/0-abstract2.tex}

\input{sections/1-intro3.tex}

\input{sections/2-background.tex}

\input{sections/3-profiling.tex}

\input{sections/3.1-design}

\input{sections/3.2-overhead}

\input{sections/4-analysis.tex}

\input{sections/4.1-model.tex}

\input{sections/4.x_layer_info.tex}

\input{sections/4.2-layer.tex}

\input{sections/4.x_layer_kernel_info.tex}

\input{sections/4.x_gpu_name_info.tex}

\input{sections/4.3-gpu.tex}

\input{sections/3.3-extensibility}

\input{sections/4.x_gpu_layer_info.tex}

\input{sections/4.x_gpu_model_info.tex}

\input{sections/5.0-systems_list.tex}

\input{sections/5.x-tf_models.tex}

\input{sections/5.xx-tf_models_cls.tex}

\input{sections/5-evaluation.tex}

\input{sections/5.xxx-mx_models_cls.tex}

\input{sections/5.1-model.tex}

\input{sections/5.2-framework.tex}

\input{sections/5.3-system.tex}

\input{sections/7-conclusion.tex}

\input{sections/99-ack.tex}

\bibliography{IEEEabrv,main}

\end{document}

%% file: sections/0-abstract2.tex
\begin{abstract}
There has been a rapid proliferation of machine learning/deep learning (ML) models and 
wide adoption of them in many application domains.
This has made profiling and characterization of ML model performance an increasingly pressing task for both hardware designers and system providers, as they would like to offer the best possible system to serve ML models with the target latency, throughput, cost, and energy requirements while maximizing resource utilization.
Such an endeavor is challenging as the characteristics of an ML model depend on the interplay between the model, framework, system libraries, and the hardware (or the HW/SW stack).
Existing profiling tools are disjoint, however, and only focus on profiling within a particular level of the stack, which limits the thoroughness and usefulness of the profiling results.

This paper proposes {\carml} --- an across-stack profiling design that gives a holistic and hierarchical view of ML model execution.
{\carml} leverages
  distributed tracing to aggregate and correlate profile data from different sources.
\carml{} introduces a leveled and iterative measurement approach that accurately captures the latencies at 
all levels of the HW/SW stack
in spite of the profiling overhead.
We couple the profiling design with an automated analysis pipeline to systematically analyze $65$ state-of-the-art ML models.
We demonstrate that {\carml} provides insights which would be difficult to discern otherwise.
\end{abstract}

%% file: sections/1-intro3.tex
\section{Introduction}\label{sec:introduction}

Machine learning/deep learning (ML) models are increasingly being used to solve problems across many domains such as image classification, object detection, machine translation, etc.
This has resulted in a surge of interest in optimizing and deploying these models on many hardware types including commodity servers, accelerators, reconfigurable hardware, mobile/edge devices, and ASICs.
As a result, there is an increasing need to profile and understand the performance of ML models.

\begin{figure}   
\centering
\includegraphics[width=0.48\textwidth]{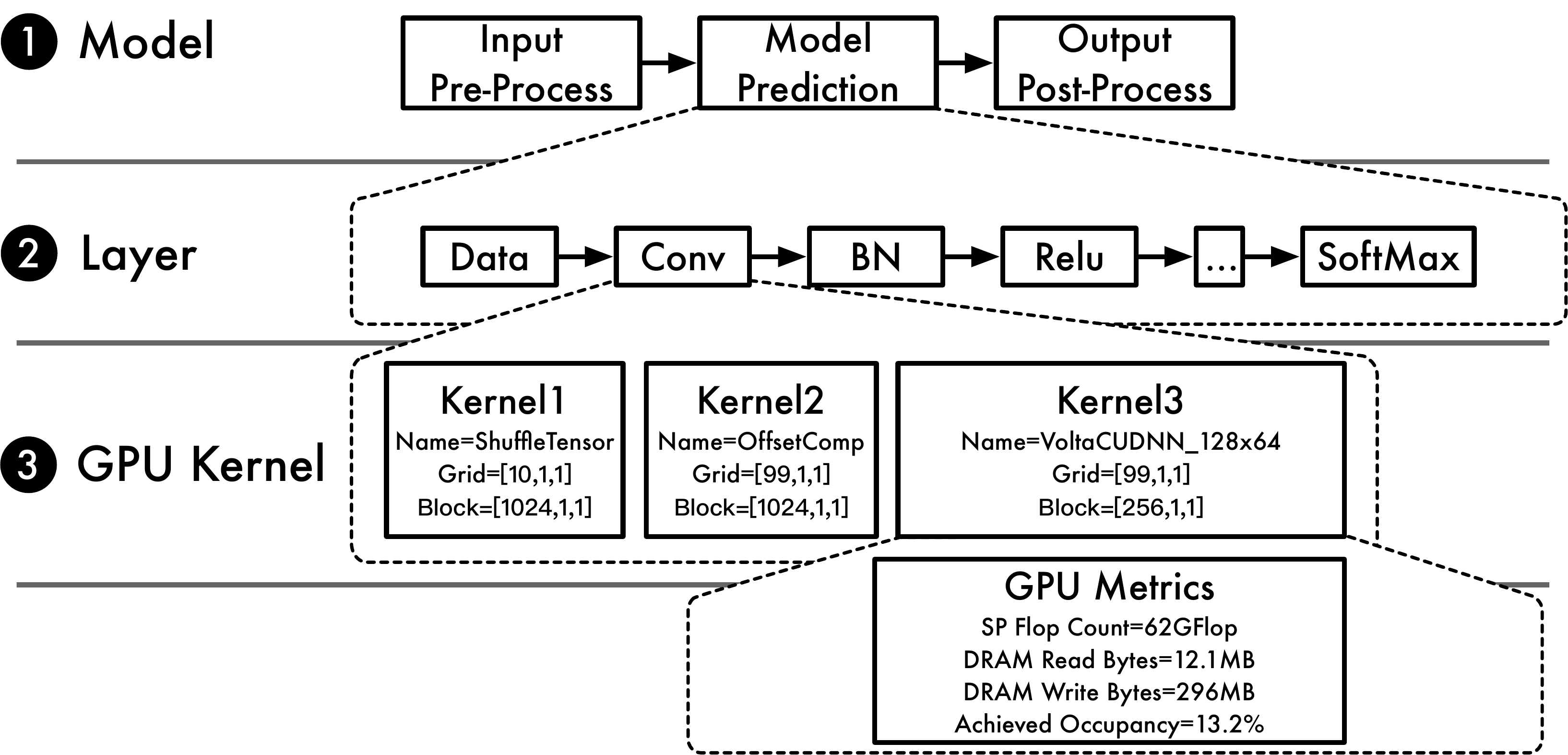}
\caption{The model-, layer-, and GPU kernel-level profile of  \texttt{MLPerf\_ResNet50\_v1.5} (Table \ref{tab:tf_models}) on \texttt{Tesla\_V100} (Table~\ref{tab:systems}) with batch size $256$ using NVIDIA GPU Cloud TensorFlow v$19.06$. 
The layers executed are data (Data), convolution (Conv), batch normalization (BN), relu (Relu), etc. 
The $3$ GPU kernels from the first Conv layer are shown along with the GPU metrics of Kernel $3$.
}
\label{fig:gpu_profile}
\end{figure}

Characterizing ML model inference is complex as its performance depends on the interplay between different levels of the HW/SW stack --- frameworks, system libraries, and hardware platforms.
Figure~\ref{fig:gpu_profile} shows an example model inference pipeline on GPUs.
At the top, there is the \circled{1} model-level evaluation pipeline.
Components at the model-level include input pre-processing, model prediction, and output post-processing.
Within the model prediction step are the \circled{2} layer-level components --- layer operators including convolution (Conv), batch normalization (BN), softmax, etc.
Within each layer are the \circled{3} GPU kernel-level components --- a sequence of CUDA API calls or GPU kernels invoked by the layer.
Because of the complexities of model inference, one needs a holistic view of the execution to identify and locate performance bottlenecks.

Existing profiling tools or methods only provide a partial view of model execution. 
To capture a holistic view of model execution, one has to switch between an array of tools.
Take the current ML profiling on GPUs for example.
To measure the model-level latency, one inserts timing code around the model prediction step of the inference pipeline.
To capture the layer-level information, one uses the ML framework's profiling capabilities~\cite{tfprofiler, mxprofiler}.
And, to capture GPU kernel information, one uses GPU profilers such as NVIDIA's nvprof~\cite{nvprof} or Nsight~\cite{nsight}.
The output profiles from the different tools are disjoint; e.g., the GPU kernels are not correlated with the layers.
As a result, one cannot construct Figure~\ref{fig:gpu_profile} and identify that the three GPU kernels shown come from the first Conv layer, for example.
This same issue exists when profiling ML model execution on CPUs. %

To correlate profiled events with model layers, vendors modify ML frameworks and instrument them to work with their profilers.
For example, NVIDIA GPU Cloud~\cite{ngc} (NGC) hosts frameworks which are instrumented with NVTX~\cite{nvtx} markers.
The NVTX markers are added around each layer in the framework and are captured along with GPU events by Nvidia's nvprof and Nsight profilers.
However, this approach only annotates GPU kernel-level information with layer names and lacks the layer-level profiling reported by the framework.
Moreover, using these instrumented frameworks creates vendor lock-in --- making the profiling and analysis dependent on the vendor's frameworks and profilers.
This is not an option for ML models developed or deployed using customized or non-vendor supported frameworks. %

To address the above issue, we propose \carml{} --- an across-stack profiling design along with a leveled experimentation methodology.
\carml innovatively leverages distributed tracing to aggregate and correlate the profiles from different sources into a single timeline trace.
Through the leveled experimentation methodology, \carml copes with the profiling overhead and accurately captures the profiles at each HW/SW stack level.
Users can use \carml to have a smooth hierarchical step-through of model performance at different levels within the HW/SW stack and identify bottlenecks.
Unlike existing approaches, \carml requires no framework modifications.
We implement the profiling design for GPUs and couple it with an across-stack analysis pipeline.
The analysis pipeline consumes the across-stack profiling trace and performs $15$ types of automated analyses (Table~\ref{tbl:analysis_list}).
These analyses allow us to characterize ML models and their interplay with frameworks, libraries, and hardware.
The consistent profiling and automated analysis workflows in \carml enable systematic comparisons of models, frameworks, and hardware.

In summary, this paper makes the following contributions:

\begin{itemize}[nosep,leftmargin=0.5em,labelwidth=*,align=left] %
	\item We propose \carml, an across-stack profiling design that innovatively leverages distributed tracing to aggregate profile data from different profiling sources and construct a holistic view of ML model execution. %
	\item We introduce a leveled experimentation methodology that allows \carml to
    accurately capture the profile at each HW/SW stack level despite the profiling overhead.
    \item We implement the design for GPU ML model inference and couple it with an analysis pipeline that performs $15$ types of automated analyses to systematically characterize ML model execution.
    \item We conduct comprehensive experiments to show the utility of \carml{}.
    We use $65$ state-of-the-art ML models from MLPerf Inference, AI-Matrix, and TensorFlow and MXNet model zoos. We evaluate the models on $5$ representative systems that span the past $4$ GPU generations (Turing, Volta, Pascal, and Maxwell) and present performance insights that would otherwise be difficult to discern absent \carml{}.
\end{itemize}

The rest of the paper is organized as follows. Section~\ref{sec:background} describes the current profiling tools and benchmarking efforts within the ML and system communities.
Section~\ref{sec:profiling} presents our design and implementation. Section~\ref{sec:analysis} showcases $15$ types of automated analysis that can be performed.
Section~\ref{sec:evaluation} further evaluates $65$ ML models and presents some insights that are enabled by our design. Section~\ref{sec:conclusion} concludes this paper.

%% file: sections/2-background.tex
\section{ ML Profiling on GPUs and Related Work}\label{sec:background}

Researchers leverage different tools and methods to profile ML model execution at each specific level of the HW/SW stack on GPUs.
Figure~\ref{fig:gpu_profile} illustrates the model-, layer-, and GPU kernel-level profiling levels on GPUs.

\circled{1} \textbf{Model-level} profiling measures the steps within the model inference pipeline.
There exist active efforts by both research and industry to develop benchmark suites~\cite{mlperf,aimatrix}  to measure and characterize models under different workload scenarios.
For model-level profiling, researchers manually insert timing code around inference steps such as input pre-processing, model prediction, and output post-processing.
Researchers then use the results as reference points to compare models or systems. %

\circled{2} \textbf{Layer-level} profiling measures the layers executed by the ML framework using the framework's profilers~\cite{tfprofiler,mxprofiler}.
These framework profilers are either built-in to the framework or are community-contributed framework plugins.
The layer index, name, latency, and memory allocations are captured by the framework profiler as it is executing the layers.
Researchers explicitly enable the framework's profiler in their code to get the layer-level profile in a framework-specific format.

\circled{3} \textbf{GPU kernel-level} profiling measures the low-level GPU information.
Using NVIDIA's nvprof and Nsight profilers, researchers capture the executed GPU kernels information such as their name, latency and metrics.
NVIDIA's nvprof and Nsight profilers are built on top of the NVIDIA CUPTI library~\cite{cupti}, which provides an API to capture CUDA API, GPU kernel, and GPU metric information.

The disconnect between the above profiling levels prohibits researchers from being able to have a holistic view of model execution --- thus, limiting the types of analysis which can be performed.
Take the \verb!MLPerf_ResNet50_v1.5! model in Figure~\ref{fig:gpu_profile} for example.
One can use the aforementioned profiling tools to get the most time-consuming layer (the $208\textsuperscript{th}$ layer which is named \texttt{conv2d\_\allowbreak 48/Conv2D}) and the most time-consuming GPU kernel (\texttt{volta\_\allowbreak scudnn\_\allowbreak 128x64\_\allowbreak relu\_\allowbreak interior\_\allowbreak nn\_\allowbreak v1}).
However, because of the lack of correlation between the GPU kernels and the layers, no other useful analysis can be performed.
E.g, %
one cannot figure out the GPU kernels invoked by the most time-consuming layer, or correlate the most time-consuming GPU kernel to a specific layer within the model.
Knowing the correlation between layers and GPU kernels enables more meaningful analyses and informs more optimization opportunities.

Currently, other than modifying framework source code, no tool or method exists to correlate the GPU kernel-level profile to the layer-level profile. 
For example, to be able to correlate GPU kernels to a certain layer, researchers manually instrument the framework's source code with NVTX markers to annotate layers~\cite{nvtxplugins}.
The NVTX markers are captured by the nvprof or Nsight profilers and kernels within the markers' ranges belong to the annotated layers.
Since the correlation between GPU kernels and layers is highly desired, NVIDIA provides modified versions of frameworks as Docker containers (NGC) where the frameworks are already instrumented with NVTX markers.
While the profile captured in this approach correlates GPU kernels with layers, it lacks critical layer-level profiling (such as memory allocations performed by a framework for a layer).
Furthermore, current implementations~\cite{nvtxplugins} introduce barriers which inhibit frameworks from performing certain optimizations (such as layer-fusion) since the NVTX layer marking is performed by surrounding each layer with a ``start NVTX marker'' layer and an ``end NVTX marker'' layer.
Finally, using vendor frameworks is not an option for profiling ML models developed with customized frameworks --- a common practice when using user-defined layers. %

To overcome the unknown correlation between layers and GPU kernels without vendor lock-in, there have been efforts~\cite{convbench,deepbench} to develop fine-grained micro-benchmarks of representative layers.
These micro-benchmarks target convolution or RNN layers and are purposely built for algorithm developers, compiler writers, and system researchers.
Using layer parameters of popular models, these micro-benchmark measure each layer in isolation.
Thus, they do not reflect how layers are executed by frameworks.
At best, micro-benchmarks give a lower-bound estimate of how layers would perform in an ideal scenario.
This lower-bound can be used to pinpoint potential optimizations in the HW/SW stack~\cite{benanza}.
Recent benchmark suites take a multi-tier approach~\cite{deep500,aimatrix} and provide a collection of benchmarks that cover both end-to-end model and layer benchmarking.

We believe a profiling design which captures ML model executions at different HW/SW stack levels and correlates profile data from the different sources --- coupled with automated analyses of the results --- would boost the productivity of researchers and help understand the model/system performance and identify the bottlenecks.
The authors are unaware of any previous work on the aforementioned across-stack profiling.
Hence, we design \carml{}.

%% file: sections/3-profiling.tex
\section{\carml{} Design and Implementation}\label{sec:profiling}

%% file: sections/3.1-design.tex
\subsection{Across-Stack Profiling Through Distributed Tracing}\label{sec:design}

To incorporate profile data from different sources and to create a holistic hierarchical view of ML model execution, \carml{}  leverages distributed tracing~\cite{tracecontext,opentracing,opentelemetry}.
This section presents \carml's across-stack profiling design.
  
Distributed tracing is a technique originally conceived for distributed applications, e.g. the ones built using a micro-service architecture.
In distributed tracing terminology, a timed operation representing a piece of work is referred to as a \textit{span}.
Each span contains a unique identifier (used as its reference), start/end timestamps, and user-defined annotations such as name, key-value tags, and logs.
A span may also contain a \textit{parent reference} to establish a parent-child relationship.
Each service in a distributed application has a \textit{tracer} --- some code to create and publish spans.
Spans are published to a \textit{tracing server} which is run on a local or remote system.
The tracing server aggregates the spans published by the different tracers into one application timeline trace.

We observe similarities between distributed tracing and across-stack profiling.
Based on this observation, we propose \carml{}, an across-stack profiling design. %
Profiling across stack levels can be represented using the distributed tracing terminology by: \circledwhite{1} each profiler within a stack is turned into a tracer, \circledwhite{2} the profiled events each form a span, \circledwhite{3} each span is tagged with its stack level, and \circledwhite{4} the parent-child relationship is encoded using a parent reference.
The conversion from the profiled events to spans can be performed online while the profiler is running, or can be performed off-line by processing the output of the profiler.
The published spans across the stack levels are aggregated by a tracing server into a single timeline trace.
Multiple tracers (or profilers) can exist within a stack level, e.g. both CPU and GPU tracers can co-exist at system library or hardware level.
As a feature supported by distributed tracing, tracers can be enabled or disabled at runtime.

During span creation, we can, in some cases, associate it with a parent (e.g. map the layer-level spans to the model prediction span).
In other cases, because of the use of disjoint profilers, manually associating the child span with its immediate parent is not possible (e.g. map the GPU kernel-level spans to the layer-level spans).
To reconstruct the missing parent-child relationship of the profiled events captured by different profilers, \carml{}'s profile analysis builds an interval tree~\cite{pal2009interval} and populates it with intervals corresponding to the spans' start/end timestamps.
Using the interval tree,  \carml{} reconstructs the parent-child relationship by checking for interval set inclusion (if the interval span $s_1$ contains the interval span $s_2$ and the level of $s_1$ is one level higher than the level of $s_2$, then $s_1$ is a parent of $s_2$).
It is possible that there are parallel events where it may be ambiguous to determine a span's parent.
In those cases, \carml{} requires another profiling run where the parallel events are serialized to get the missing correlation information.
This can be performed by specifying environment variables without modifications to the application  --- e.g. setting either \texttt{CUDA\_LAUNCH\_BLOCKING=1} for GPUs using CUDA or \texttt{OMP\_NUM\_THREADS=1} for CPUs using OpenMP.

To profile asynchronous functions, \carml{} captures two spans for each asynchronous function denoting their asynchronous launch (called a \textit{launch span}) and future execution (called an \textit{execution span}).
\carml{} correlates the two spans using a correlation identifier which is inserted as a span tag during span creation.
\carml{} uses the launch span's parent as the parent of the asynchronous function and uses the execution span to get the performance information or find child spans. 
E.g., to profile asynchronous GPU kernels, \carml{} captures both the kernel launch and execution spans (as detailed in Section~\ref{sec:xspgpu}).

\begin{figure} 
\centering
\includegraphics[width=0.48\textwidth]{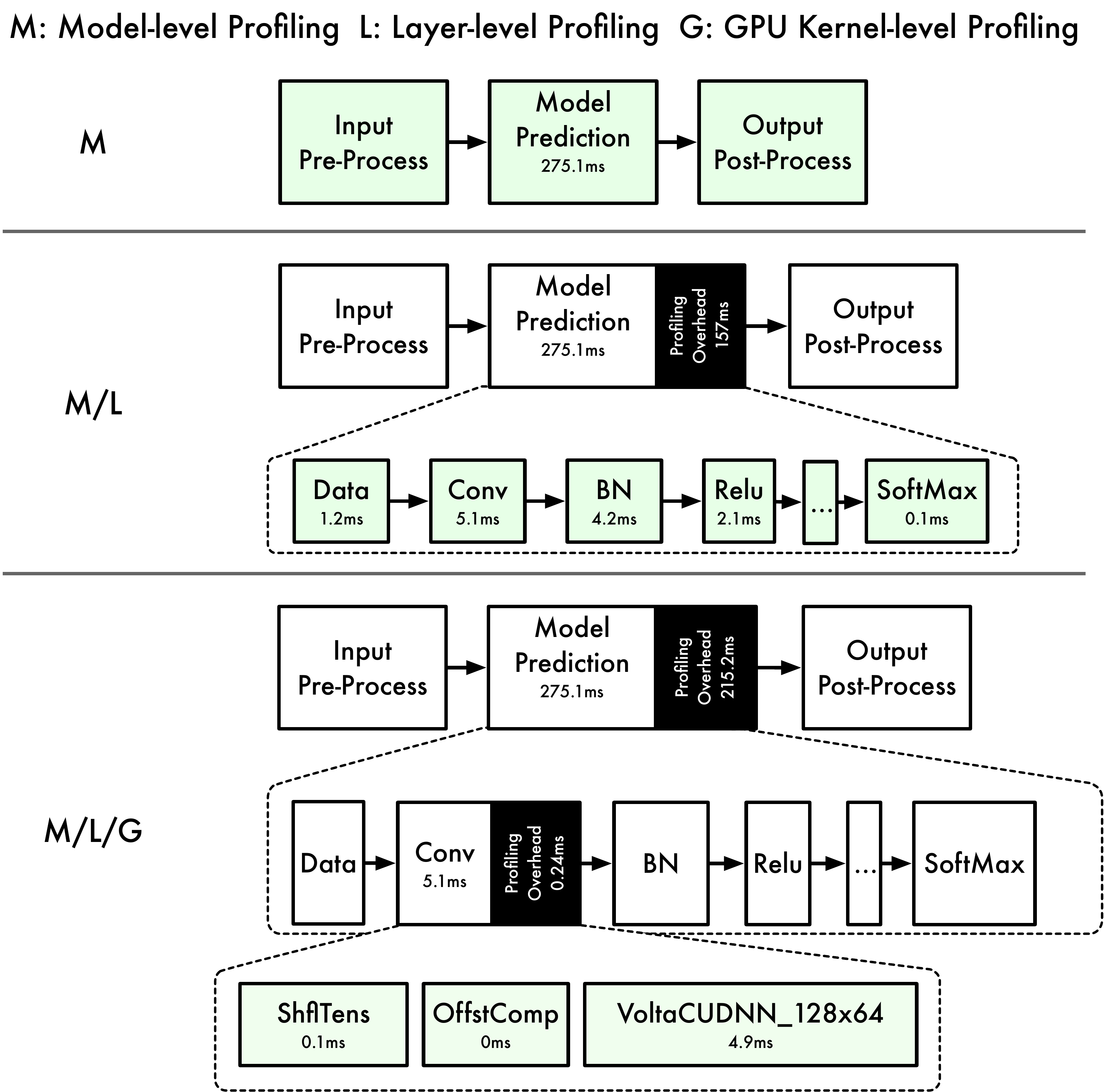}
\caption{\carml{} profiles for \texttt{MLPerf\_ResNet50\_v1.5} with batch size $256$ on \texttt{Tesla\_V100} (Table VI) with the model-level (M), model-/layer-level (M/L), and model-/layer-/GPU kernel-level (M/L/G) profiling. At each level, the green components correctly measure the latency whereas the rest incur profiling overhead. }
\label{fig:profiling_levels_overheads}
\end{figure}

\subsection{Across-stack Profiling on GPUs}\label{sec:xspgpu}

While the across-stack profiling design presented above is general, this paper focuses on the profiling of ML models on GPUs across the model, layer, and GPU kernel level:

\circled{1} \textbf{Model-level profiling} ---
To profile at the model granularity, \carml{} provides tracing APIs --- \texttt{startSpan} and \texttt{finishSpan} --- which can be placed within the inference code to measure code regions of interest.
For example, to measure the time spent running the model prediction using the framework C APIs, one places the tracing APIs around the calls to \texttt{TF\_\allowbreak SessionRun} for TensorFlow or \texttt{MXPredForward} for MXNet.
This only requires adding two extra lines in the user's inference code.

\circled{2} \textbf{Layer-level profiling} --- 
To profile at the layer granularity, \carml{} uses the ML framework's existing profiling capability.
During runtime, \carml{} enables the framework profiler, converts the profile results into spans, and publishes them to the tracing server.
In TensorFlow, enabling layer profiling requires calling the framework's prediction function with the profiling option enabled.
This option is controlled by the \texttt{RunOptions.TraceLevel} setting which is passed to the \texttt{TF\_SessionRun} function in TensorFlow.
In MXNet, the \texttt{MXSetProfilerState} function enables and disables layer profiling.
Similar mechanisms exist for other frameworks such as Caffe, Caffe2, PyTorch, and TensorRT.
The layer spans are set to be the children of the model prediction span, and hence each layer are directly correlated to the model prediction step.
Since \carml{} leverages the existing framework's profiling capabilities, profiling at the layer level require no modification to the framework's source code. %

\circled{3} \textbf{GPU kernel-level profiling} --- 
To obtain the GPU profile, \carml{} uses NVIDIA's CUPTI library~\cite{cupti}.
The CUPTI library captures the CUDA API calls, GPU activities (GPU tasks such as kernel executions and memory copies), and GPU kernel metrics (low-level hardware counters such as GPU achieved occupancy, flop count, and memory read/write for GPU kernels).
Similar to Nsight or nvprof (which are built on top of CUPTI), one can specify with \carml{} which CUDA APIs, GPU activities, or metrics to capture.
At runtime, \carml{} converts the captured CUPTI information  into spans and publishes them to the tracer server (asynchronously to avoid added overhead).
If profiling GPU metrics is enabled, the metrics are added as metadata to the corresponding kernel's span.

GPU kernels are often launched asynchronously by the ML frameworks or libraries.
Therefore, for each kernel two spans are created within the \carml{} timeline.
\carml{} uses the CUPTI callback API to capture the CUDA API call \texttt{cudaLaunchKernel} as the launch span.
\carml{} uses the CUPTI activity API to capture the effective kernel duration as the execution span.
\carml{} uses the kernel launch span to associate it with the parent layer span and use the execution span to get the kernel performance information.
The two spans are correlated by the \texttt{correlation\_id} provided by CUPTI.
Since this correlation can potentially be expensive, we perform correlation during profile analysis which aggregates the information from two GPU kernel spans.

%% file: sections/3.2-overhead.tex
\subsection{Dealing with Profiling Overhead through Leveled Experimentation}

Profiling always comes with overhead.
We observe that creating spans online adds negligible overhead per span (and no overhead exists if the profile is converted offline).
Thus, \carml{} incurs only the profiling overhead introduced by the integrated profilers.
For example, layer-level profiling adds overhead to the model prediction depending on how many layers are executed.
And as with the existing NVIDIA profilers, the GPU-level profiling incurs overhead, which can be substantial depending on if GPU metric profiling is enabled and the types of GPU metrics to capture.
GPU memory metrics are especially expensive to profile and can slow down execution by over $100\times$.
This is due to the limited number of GPU hardware performance counters, which require GPU kernels to be replayed multiple times to capture the user-specified metrics.

Profilers at a specific stack level accurately capture the events within that level.
And, since tracers in \carml can be enabled or disabled depending on the characterization target, the profiling overhead can be controlled by picking the profiling level.
For an event at level $n$ (where level $1$ is the model level), the profiling overhead introduced at level $n+1$ can be quantified by subtracting the latency of the event when profilers up to level $n$ are enabled from the latency when profilers up to level $n+1$ are enabled.
We refer to the profiling practice which uses traces from multiple runs with different profiling levels as \textit{leveled experimentation}.
Through leveled experimentation, \carml{} gets accurate timing of profiled events at all stack levels.

To demonstrate the profiling overhead and the leveled experimentation, we use the \texttt{MLPerf\_\allowbreak ResNet\allowbreak 50\_\allowbreak v1.5} model running on the \texttt{Tesla\_V100} system (Table~\ref{tab:systems}) as an example.
Figure~\ref{fig:profiling_levels_overheads} shows the model’s
\carml{} profiles at different profiling levels.
We can enable the model-level profiling (M) to get the baseline model prediction latency of $275.1 ms$.
To further measure the latency of each layer, we enable both the model- and layer-level profiling (M/L).
While the layer-level profiling adds overhead to the model prediction latency, it accurately captures the latency of each layer.
We can quantify this overhead by subtracting the model prediction latency in the model-level profile from the model prediction latency in the model-/layer-level profile.
We find that the layer-level profiling introduces a $157ms$ overhead.
We can further perform the GPU kernel-level profiling along with the model-/layer-level profiling to get a hierarchical view of the model execution (M/L/G).
Enabling the GPU kernel-level profiling adds extra overhead to the model prediction latency --- making the model prediction step (with the added overhead) take $490.3ms$.
If we look at the first convolution layer, the GPU profiling of the $3$ child kernels incurs a $0.24ms$ overhead.
We verified the layer and GPU kernel latencies measured by \carml{} against what framework and NVIDIA’s profilers report.

%% file: sections/4-analysis.tex
\input{sections/xx-analysis-table.tex}

\subsection{Across-Stack Analysis}\label{sec:analysis}

We couple \carml{} with an automated analysis pipeline which consumes the profiling traces published to the tracing server.
We define $15$ analyses that capture across-stack characteristics of ML model executions on GPUs as listed in Table~\ref{tbl:analysis_list}.
The $15$ analyses are grouped into $3$ categories based on the profiling information required.
Since meaningful characterization requires multiple runs, the pipeline takes traces from a user-defined number of evaluations, correlates the information, and computes the trimmed mean value (or other user-defined statistical summaries) for the same performance value (e.g. latency) across runs.
This automated analysis pipeline allows users to systematically and efficiently characterize and compare ML models.

To illustrate the analyses, we use the TensorFlow \texttt{MLPerf\_ResNet50\_v1.5} model (ID $= 7$ in Table~\ref{tab:tf_models}) from the MLPerf Inference v$0.5$ release.
The model is run within the NGC TensorFlow container v$19.06$ on an AWS P3~\cite{awsp3} instance (\texttt{Tesla\_V100} in Table~\ref{tab:systems}).
The P3 instance is equipped with a Tesla V100-SXM2 GPU and achieves a peak throughput of $15.7$ TFlops and $900$ GB/s global memory bandwidth.
Batch size $256$ is used in Sections \ref{sec:layerlvl} and \ref{sec:gpulvl}, since the model achieves maximum throughput at that batch size.
Using \carml, one can perform analyses that are either difficult or impossible using existing tools or methods.

%% file: sections/xx-analysis-table.tex
\begin{table}
    \centering
    \caption{The $15$ analyses performed by \carml. The analyses require profiling information from one or more levels (\textbf{M}: model-level, \textbf{L}: layer-level, and \textbf{G}: GPU kernel-level).}
    \resizebox{0.48\textwidth}{!}{%
    \begin{tabular}{rlcccccc}
        \hline
          & \textbf{Analysis} & \shortstack{\textbf{Profiling} \\ \textbf{Levels}}  & \shortstack{\textbf{End-to-End} \\ \textbf{Benchmarking}}  & \shortstack{\textbf{Framework} \\ \textbf{Profilers}}   & \shortstack{\textbf{NVIDIA} \\ \textbf{Profilers}}  & \textbf{\carml} \\
        \hline
        \analysis{1} & Model information table & \textbf{M} & \yes & \no & \no & \yes  \\
        \analysis{2} & Layer information table & \textbf{L} & \no & \yes & \no & \yes \\
        \analysis{3} & Layer latency & \textbf{L} & \no & \yes & \no & \yes  \\
        \analysis{4} & Layer memory allocation & \textbf{L} & \no & \yes & \no & \yes  \\
        \analysis{5} & Layer type distribution & \textbf{L} & \no & \yes & \no & \yes  \\
        \analysis{6} & Layer latency aggregated by type & \textbf{L} & \no & \yes & \no & \yes  \\
        \analysis{7} & Layer memory allocation aggregated by type & \textbf{L} & \no & \yes & \no & \yes  \\
        \analysis{8} & GPU kernel information table & \textbf{G} & \no & \no & \yes & \yes  \\
        \analysis{9} & GPU kernel roofline & \textbf{G} & \no & \no & \yes & \yes  \\
        \analysis{10} & GPU kernel information aggregated by name table & \textbf{G} & \no & \no & \yes & \yes  \\
        \analysis{11} & GPU kernel information aggregated by layer table & \textbf{L/G} & \no & \no & \no & \yes  \\
        \analysis{12} & GPU metrics aggregated by layer & \textbf{L/G} & \no & \no & \no & \yes \\
        \analysis{13} & GPU vs Non-GPU latency & \textbf{L/G} & \no & \no & \no & \yes \\
        \analysis{14} & Layer roofline & \textbf{L/G} & \no & \no & \no & \yes \\
        \analysis{15} & GPU kernel information aggregated by model table & \textbf{M/G} & \no & \no & \yes & \yes \\
        \hline
    \end{tabular}	}%
    \label{tbl:analysis_list}
\end{table}

%% file: sections/4.1-model.tex
\subsubsection{Using Model-level Profile}\label{sec:modellvl}

\begin{figure}
\centering
\includegraphics[width=0.48\textwidth]{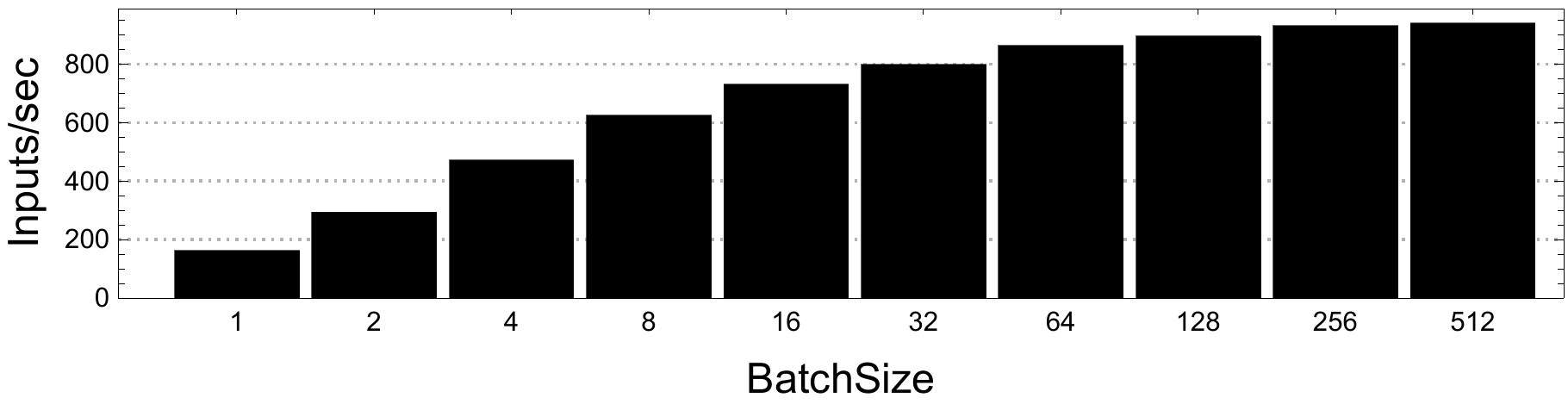}
\caption{The throughput of \texttt{MLPerf\_ResNet50\_v1.5} across batch sizes on \texttt{Tesla\_V100}.}
\label{fig:model_throughput}
\end{figure}

Both model throughput and latency are important to researchers who want to understand a model's end-to-end performance.
Using only the model-level profiling, \carml{} automates the computation of a model's throughput and latency across batch sizes and generate a \analysis{1} model information table.
\carml{} then computes the model's optimal batch size given a user-defined metric (e.g. a latency target). 
By default \carml{} computes the optimal batch size by evaluating the model across batch sizes and selecting the batch size where doubling it does not increase the model's throughput by more than $5\%$.
Figure~\ref{fig:model_throughput} shows the throughput of \texttt{MLPerf\_\allowbreak ResNet50\_\allowbreak v1.5} across batch sizes.
\carml{} computes the optimal batch size as $256$ where the model achieves a maximum throughput of $930.7$ images/second.
The corresponding batch latency is $275.05ms$.
Absent \carml{}, researchers insert timing functions around the model prediction code, perform multiple evaluations, and write scripts to compute the model's throughput, latency, and optimal batch size.

%% file: sections/4.x_layer_info.tex
\begin{table}
    \centering
    \caption{The top $5$ most time consuming layers in \analysis{2}~ for \texttt{MLPerf\_ResNet50\_v1.5} with batch size $256$ on \texttt{Tesla\_V100}. In total, there are $234$ layers of which $143$ take less than $1$ ms.}
    \resizebox{0.48\textwidth}{!}{%
        \begin{tabular}{rlccrr} \toprule
        \centering%
        \shortstack{\textbf{Layer} \\ \textbf{Index}} & \shortstack{\textbf{Layer} \\ \textbf{Name}} & \shortstack{\textbf{Layer} \\ \textbf{Type}} & \shortstack{\textbf{Layer} \\ \textbf{Shape}}  & \shortstack{\textbf{Latency} \\ \textbf{(ms)}} & \shortstack{\textbf{Alloc Mem} \\ \textbf{(MB)}} \\   \midrule
        208 & \texttt{conv2d\_48/Conv2D}	& Conv2D & $\langle 256, 512, 7, 7 \rangle$ & 7.59 & 25.7 \\
        221 & \texttt{conv2d\_51/Conv2D} & Conv2D & $\langle 256, 512, 7, 7 \rangle$ & 7.57 & 25.7 \\
        195 & \texttt{conv2d\_45/Conv2D}	& Conv2D & $\langle 256, 512, 7, 7 \rangle$ & 5.67 & 25.7 \\
        3 & \texttt{conv2d/Conv2D} & Conv2D & $\langle 256, 64, 112, 112 \rangle$ & 5.08 & 822.1 \\
        113 & \texttt{conv2d\_26/Conv2D}	& Conv2D & $\langle 256, 256, 14, 14 \rangle$ & 4.67 & 51.4 \\
        \bottomrule
        \end{tabular}%
	}
    \label{tab:layer_info}
\end{table}

%% file: sections/4.2-layer.tex
\begin{figure}
    \centering
    \setlength{\abovecaptionskip}{0pt}
    \includegraphics[width=0.48\textwidth]{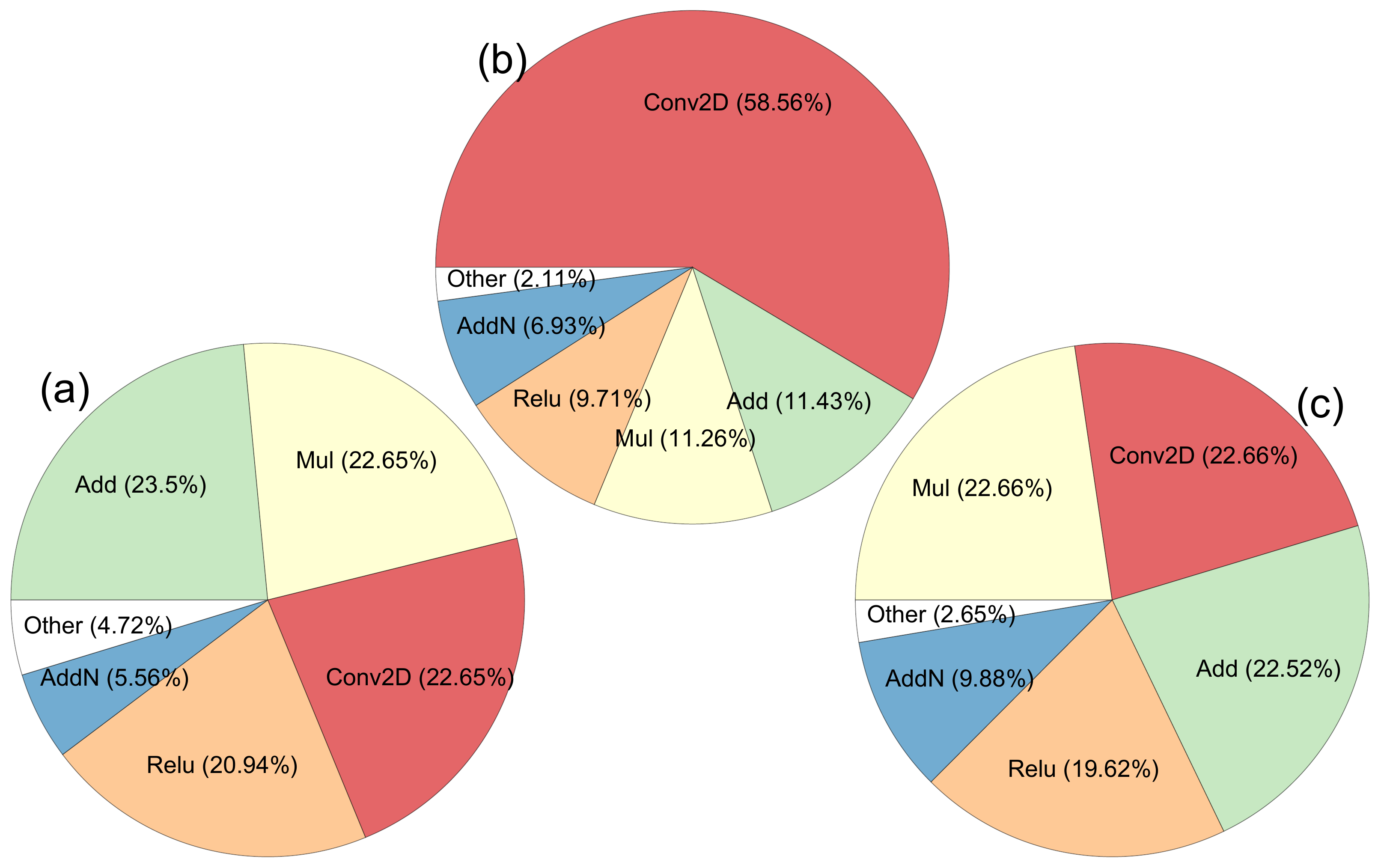}
        \caption{Layer statistics for \texttt{MLPerf\_ResNet50\_v1.5} on \texttt{Tesla\_V100}: (a) \analysis{5}~ layer type distribution, (b) \analysis{6}~ layer latency aggregated by type, (c) \analysis{7}~ layer memory allocation aggregated by type.}
    \label{fig:layer_infos}
\end{figure}

\begin{figure*}
    \centering
    \setlength{\abovecaptionskip}{0pt}
    \includegraphics[width=0.98\textwidth]{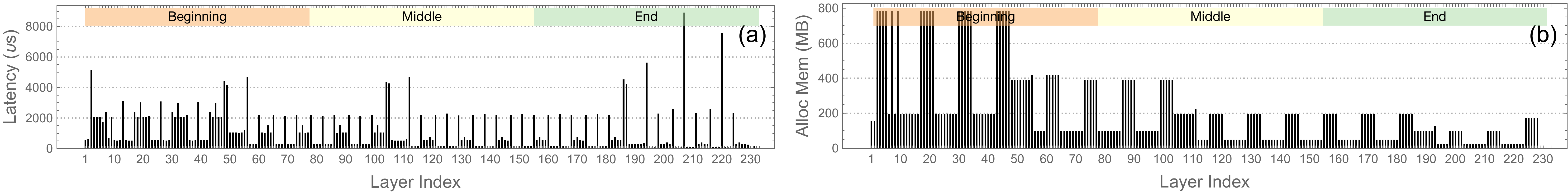}
    \caption{The (a) \analysis{3}~ latency and (b) \analysis{4}~ memory allocation for each layer in \texttt{MLPerf\_ResNet50\_v1.5} with batch size $256$ on \texttt{Tesla\_V100}.
    To understand the performance trend, we divide the model execution into $3$ intervals based on the layer index: beginning, middle, and end.}
    \label{fig:layer_info_combined}
  \vspace{-10pt}
\end{figure*}

\subsubsection{Using Model- and Layer-level Profiles}\label{sec:layerlvl}

Using both the model- and layer-level profiles enables characterization of layers executed by the ML framework.
The measured layers may be different from the ones statically defined in the model graph, since a framework may perform model optimization at runtime.
Using the data captured, \carml{} generates a \analysis{2} layer information table reporting index, name, shape, latency, and allocated memory of all the layers.
For example, Table~\ref{tab:layer_info} shows the top $5$ most time-consuming layers for \texttt{MLPerf\_ResNet50\_v1.5}.

\carml{} further uses the profile data to visualize both the \analysis{3} latency per layer and \analysis{4} allocated memory per layer in layer execution order.
Figures~\ref{fig:layer_info_combined} shows the two analyses for \texttt{MLPerf\_ResNet50\_v1.5} at the optimal batch size. 
We observe that a layer latency and memory allocation trend exists --- the model latency can be mostly attributed to the early executed layers.
Similarly, the memory allocation is high for the early stage of the model execution, and less so during the middle and end stages.d

We can group the layer information by layer type to derive useful layer execution statistics such as \analysis{5} the number of times each layer type is executed (Figure~\ref{fig:layer_infos}a),
the \analysis{6} layer latency aggregated by type (Figure~\ref{fig:layer_infos}b), and 
the \analysis{7} layer memory allocation aggregated by type (Figure~\ref{fig:layer_infos}c).
We observe that \texttt{MLPerf\_ResNet50\_v1.5} mostly comprises of \texttt{Add}, \texttt{Conv2D}, \texttt{Mul}, and \texttt{Relu} layers.
This is because of the ResNet modules which have the pattern of \texttt{Conv} $\rightarrow$ \texttt{BN} $\rightarrow$ \texttt{Relu}.
The ResNet modules get executed by TensorFlow as a \texttt{Conv2D} $\rightarrow$ \texttt{Mul} $\rightarrow$ \texttt{Add} $\rightarrow$ \texttt{Relu} layer sequence.
This same group of layers dominates both latency and memory allocation, with \texttt{Conv2D} being the most time-consuming layer type.

Absent \carml{}, researchers use the framework profiler to gather layer-level information.
Through manually parsing and aggregating the profiling output across runs, researchers can perform \analysis{2-7}. 
However, since the output format of a framework profiler is framework-dependent, the analysis scripts developed in this case are also framework-specific.

%% file: sections/4.x_layer_kernel_info.tex
\begin{table*}
    \centering
    \caption{The top $5$ most time-consuming kernels in \analysis{8}~ for \texttt{MLPerf\_ResNet50\_v1.5} on \texttt{Tesla\_V100}. In total, $375$ kernels are invoked of which $284$ take less than $1 ms$. 
    }
    \resizebox{0.85\textwidth}{!}{%
        \begin{tabular}{lrccrrrrrc} \toprule
        \centering%
         \textbf{Kernel Name} & \shortstack{\textbf{Layer} \\ \textbf{Index}} & \shortstack{\textbf{Layer} \\ \textbf{Kernel} \\ \textbf{Latency} \\ \textbf{(ms)}} & \shortstack{\textbf{Kernel} \\ \textbf{Gflops}}  & \shortstack{\textbf{Kernel} \\ \textbf{DRAM} \\ \textbf{Reads} \\ \textbf{(MB)}} & \shortstack{\textbf{Kernel} \\ \textbf{DRAM} \\ \textbf{Writes} \\ \textbf{(MB)}} & \shortstack{\textbf{Kernel} \\ \textbf{Achieved} \\ \textbf{Occupancy} \\ \textbf{(\%)}}  & \shortstack{\textbf{Kernel} \\ \textbf{Arithmetic} \\ \textbf{Intensity} \\ \textbf{(flops/byte)}}  & \shortstack{\textbf{Kernel} \\ \textbf{Arithmetic} \\ \textbf{Throughput}\\ \textbf{(Tflops/s)}} & \shortstack{\textbf{Memory} \\ \textbf{Bound?}}
        \\   \midrule
\texttt{volta\_cgemm\_32x32\_tn} & 221 & 6.04 & 77.42 & 40.33 & 43.86 & 12.18 & 876.97 & 12.82 & \no \\
\texttt{volta\_cgemm\_32x32\_tn} & 208 & 6.03 & 77.42 & 43.93 & 43.81 & 12.19 & 841.59 & 12.83 & \no \\
\texttt{volta\_scudnn\_128x128\_relu\_interior\_nn\_v1} & 195 & 5.48 & 59.20 & 27.71 & 8.40 & 15.49 & 1,563.30 & 10.80 & \no \\
\texttt{volta\_scudnn\_128x64\_relu\_interior\_nn\_v1} & 3 & 4.91 & 62.89 & 11.55 & 283.05 & 13.20 & 203.58 & 12.81 & \no \\
\texttt{volta\_scudnn\_128x128\_relu\_interior\_nn\_v1} & 57 & 4.56 & 59.24 & 34.83 & 37.64 & 15.15 & 779.55 & 12.99 & \no \\
        \bottomrule
        \end{tabular}%
	}
    \label{tab:layer_kernel_info}
    \vspace{-5pt}
\end{table*}

%% file: sections/4.x_gpu_name_info.tex
\begin{table*}
    \centering
    \caption{The top $5$ most time-consuming kernels in \analysis{10}~ for \texttt{MLPerf\_ResNet50\_v1.5} on \texttt{Tesla\_V100}. $30$ unique kernels are invoked in total.
    }
    \resizebox{0.98\textwidth}{!}{%
        \begin{tabular}{lrccrrrrrrc} \toprule
        \centering%
         \textbf{Kernel Name} &
         \shortstack{\textbf{Kernel} \\ \textbf{Count}} & \shortstack{\textbf{Kernel} \\ \textbf{Latency} \\ \textbf{(ms)}} & 
         \shortstack{\textbf{Kernel} \\ \textbf{Latency} \\\textbf{Percentage}} &
         \shortstack{\textbf{Kernel} \\ \textbf{Gflops}}  & \shortstack{\textbf{Kernel} \\ \textbf{DRAM} \\ \textbf{Reads} \\ \textbf{(MB)}} & \shortstack{\textbf{Kernel} \\ \textbf{DRAM} \\ \textbf{Writes} \\ \textbf{(MB)}} & \shortstack{\textbf{Kernel} \\ \textbf{Achieved} \\ \textbf{Occupancy} \\ \textbf{(\%)}}  & \shortstack{\textbf{Kernel} \\ \textbf{Arithmetic} \\ \textbf{Intensity} \\ \textbf{(flops/byte)}}  & \shortstack{\textbf{Kernel} \\ \textbf{Arithmetic} \\ \textbf{Throughput}\\ \textbf{(Tflops/s)}} & \shortstack{\textbf{Memory} \\ \textbf{Bound?}}
        \\   \midrule
\texttt{volta\_scudnn\_128x64\_relu\_interior\_nn\_v1} & 34 & 84.95 & 30.87 & 1,053.63 & 4,429.64 & 5,494.22 & 22.58 & 101.25 & 12,40 & \no \\
\texttt{Eigen::TensorCwiseBinaryOp$<$scalar\_product\_op$>$} & 52 & 28.43 & 10.33 & 2.85 & 4,181.23 & 6,371.12 & 49.72 & 0.26 & 0.10 & \yes \\
\texttt{Eigen::TensorCwiseBinaryOp$<$scalar\_sum\_op$>$} & 51 & 26.38 & 9.59 & 2.64 & 4,063.49 & 6,052.22 & 49.69 & 0.25 & 0.10 & \yes \\
\texttt{Eigen::TensorCwiseBinaryOp$<$scalar\_max\_op$>$} & 48 & 24.71 & 8.98 & 0 & 3,773.84 & 5,699.95 & 98.39 & 0 & 0 & \yes \\
\texttt{volta\_scudnn\_128x128\_relu\_interior\_nn\_v1} & 4 & 23.02 & 8.37 & 276.64 & 671.68 & 335.01 & 15.96 & 262.08 & 12,02 & \no \\
        \bottomrule
        \end{tabular}%
	}
    \label{tab:gpu_name_info}
\end{table*}

%% file: sections/4.3-gpu.tex
\subsubsection{Using Model-, Layer-, and GPU Kernel-level Profiles}\label{sec:gpulvl}

To distill fine-grained performance information, \carml{} uses  model-, layer- and GPU kernel-level profiles to 
generate a \analysis{8} GPU kernel information table summarizing all the kernels in the model prediction. 
An example is shown in Table~\ref{tab:layer_kernel_info} where the top $5$ most time consuming GPU kernel calls for \texttt{MLPerf\_ResNet50\_v1.5} are listed.
The $5$ kernels perform either matrix multiplication or convolution.
All the GPU metrics supported by the NVIDIA profiling tools~\cite{gpumetrics} can be captured through \carml{}, here we focus on \texttt{flop\_count\_sp}, \texttt{dram\_read\_bytes}, \texttt{dram\_write\_bytes}, and \texttt{achieved\_occupancy}:
\begin{itemize}[nosep,leftmargin=0.5em,labelwidth=*,align=left]
    \item \texttt{flop\_count\_sp} ---  the total number of single-precision floating-point operations executed by a kernel.
    \item \texttt{dram\_read\_bytes} --- the total number of bytes read from the GPU's DRAM to its L2 cache in a kernel.
    \item \texttt{dram\_write\_bytes} --- the total number of bytes written from the GPU's L2 cache to its DRAM in a kernel.
    \item \texttt{achieved\_occupancy} --- the ratio of the average active warps per active cycle to the maximum number of warps per streaming multiprocessor. The \texttt{achieved\_occupancy} is an indicator to the level of parallelism for a kernel.
\end{itemize}

\begin{figure}
  \centering
  \includegraphics[width=0.48\textwidth]{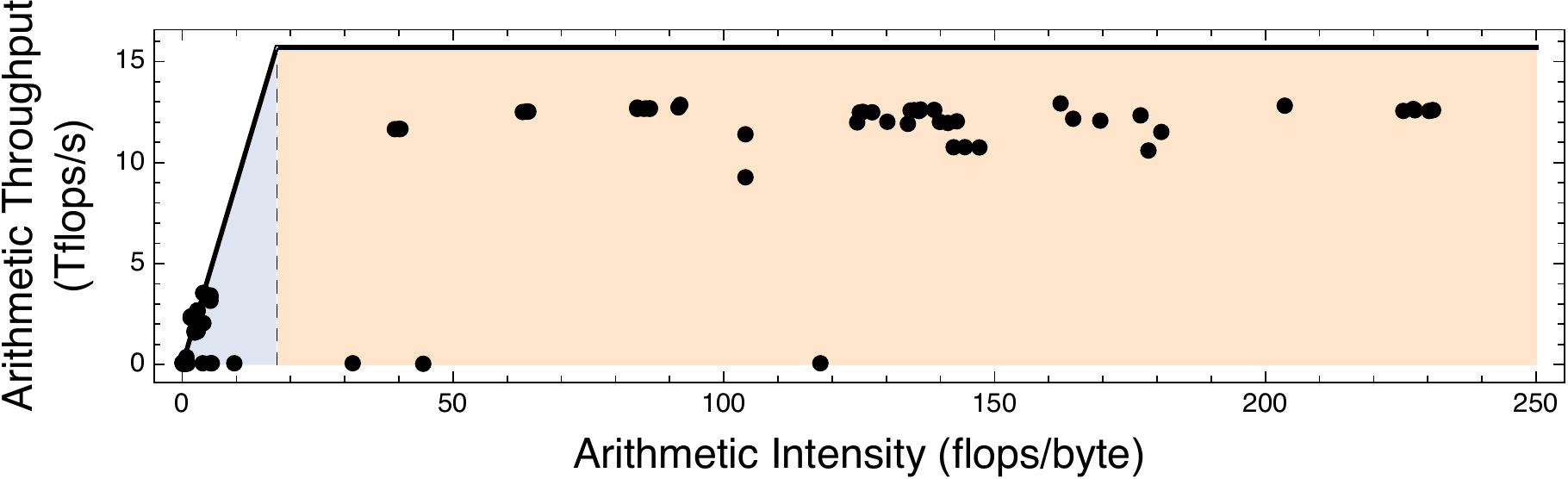}
  \caption{The \analysis{9} roofline analysis for the GPU kernels in \texttt{MLPerf\_ResNet50\_v1.5} with batch size $256$ on \texttt{Tesla\_V100}.
  Kernels within the blue region are memory-bound, whereas the ones within the orange region are compute-bound.}
  \label{fig:kernel_roofline}
\end{figure}

Using both the kernel flop and memory access metrics, \carml{} calculates the kernel arithmetic intensity and arithmetic throughput.
These parameters are used to perform GPU kernel roofline~\cite{williams2009roofline} analysis.
A kernel's arithmetic intensity is the ratio between the number of flops and the number of memory accesses:
${\small\texttt{arithmetic\_intensity} = \frac{\texttt{flop\_count\_sp}}{\texttt{dram\_read\_bytes} + \texttt{dram\_write\_bytes}}}$.
A kernel's arithmetic throughput is the ratio between the number of flops and the latency:
${\small\texttt{arithmetic\_throughput} = \frac{\texttt{flop\_count\_sp}}{\texttt{kernel\ latency}}}$.
Using the GPU's theoretical FLOPS and memory bandwidth, we compute the ideal arithmetic intensity using the equation:
${\small\texttt{ideal\_arithmetic\_intensity} = \frac{\texttt{peak\_FLOPS}}{\texttt{memory\_bandwidth}}}$.
The \texttt{Tesla\_V100} GPU, for example, has a peak throughput of $15.7$ TFLOPS and a global memory bandwidth of $900$ GB/s, hence an ideal arithmetic intensity of $\frac{15.7\ TFLOPS}{900\ GB/s} = 17.44$ flops/byte. 
A kernel is \textit{memory-bound} if its  arithmetic intensity is less than the GPU's ideal arithmetic intensity (blue region) and is \textit{compute-bound} otherwise  (orange region).
\analysis{9} visualizes the roofline analysis of all the GPU kernels (shown in Figure~\ref{fig:kernel_roofline}).
As expected, the most time-consuming kernels are convolution kernels which are all compute-bound.

\carml{} creates a table of \analysis{10} GPU kernel information aggregated by name, as shown in Table~\ref{tab:gpu_name_info}.
The aggregated kernel latency, flops, and DRAM reads and writes are calculated as the sum of all the kernel instances with the same name.
The aggregated kernel achieved occupancy is calculated as the weighted sum (by kernel latency) of achieved occupancy of all the kernel instances with the same name.
The aggregated kernel arithmetic intensity and throughput are calculated using the aggregated flops and memory accesses.
For \texttt{MLPerf\_ResNet50\_v1.5}, we observe that the most time consuming GPU kernel is \texttt{volta\_scudnn\_128$\times$64\_relu\_interior\_nn\_v1} from the cuDNN~\cite{cudnn} library, which is compute-bound and takes $30.87\%$ of the overall model prediction latency.
The $2\textsuperscript{nd}$ and $3\textsuperscript{rd}$ most time consuming kernels are \texttt{scalar\_product\_op} and \texttt{scalar\_sum\_op} and are defined by the Eigen~\cite{eigenweb} library, are memory-bound, and take $10.33\%$ and $9.59\%$ of the model inference latency, respectively.

Since each GPU kernel can be correlated to the layer that invokes it, \carml{} aggregates the information of GPU kernels within each layer and builds a table of \analysis{11} GPU kernel information aggregated by layer.
A layer's kernel latency, flops, DRAM reads and writes are calculated by adding the corresponding values of all the kernels invoked by that layer.
The layer's achieved occupancy is calculated as the weighted sum (by kernel latency) of the achieved occupancy of all the kernels within the layer.
As an example, Table~\ref{tab:gpu_layer_info} shows the aggregated GPU kernel information for the top $5$ most time-consuming layers in \texttt{MLPerf\_ResNet50\_v1.5}.

\begin{figure*}
    \centering
    \setlength{\abovecaptionskip}{0pt}
    \includegraphics[width=0.98\textwidth]{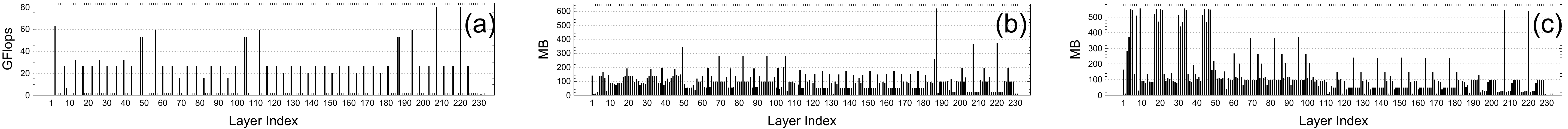}
    \caption{The \analysis{12} total GPU kernel (a) flops, (b) DRAM reads, and (c) DRAM writes per layer for \texttt{MLPerf\_ResNet50\_v1.5} with batch size $256$ on \texttt{Tesla\_V100}.}
    \label{fig:gpu_kernel_info_combined}
  \vspace{-10pt}
\end{figure*}

\begin{figure}
\centering
\includegraphics[width=0.48\textwidth]{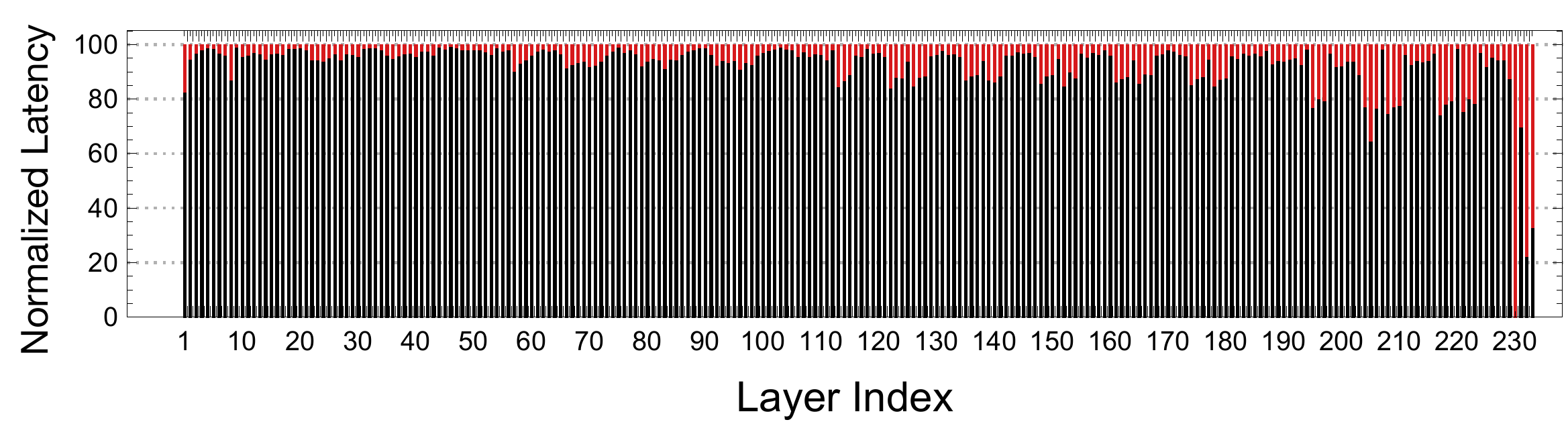}
\caption{The \analysis{13} normalized GPU and Non-GPU latency per layer for \texttt{MLPerf\_ResNet50\_v1.5} with batch size $256$ on \texttt{Tesla\_V100}. }
\label{fig:gpu_cpu_duration_per_layer}
\end{figure}

Using this data, \carml{} visualizes the \analysis{12} total flops, DRAM reads and writes per layer (shown in Figure~\ref{fig:gpu_kernel_info_combined} (a), (b) and (c) respectively). 
Subtracting a layer's total GPU kernel latency from the its overall latency computes the \analysis{13} time not spent performing GPU computation.
We call this difference the layer's \textit{non-GPU latency}.
Figure~\ref{fig:gpu_cpu_duration_per_layer} shows the layer's GPU and non-GPU latency normalized to the overall layer latency for \texttt{MLPerf\_ResNet50\_v1.5}.
The layer arithmetic intensity and throughput are calculated using the layer's total flops and memory accesses.
A \analysis{14} roofline analysis of all the layers is performed in Figure~\ref{fig:layer_roofline}. 
We observe that the \texttt{Conv2D} layers are the most compute and memory intensive. 
The \texttt{Conv2D}, \texttt{MatMul}, \texttt{BiasAdd}, and \texttt{Softmax} layers are compute-bound, whereas the other layers (\texttt{Add}, \texttt{Mul}, and \texttt{Relu}) are memory-bound.

\carml{} aggregates all the GPU kernel information within a model and computes a table of the \analysis{15} total GPU kernel latency, flop, and memory access information for the model (shown in Table~\ref{tab:gpu_model_info}). 
Similar to the layer aggregation, the model kernel latency, flops, DRAM reads and writes are calculated as the sum of all kernels invoked by the model.
\carml{} computes the model's achieved occupancy as the weighted sum (by kernel latency) of the achieved occupancy of all the kernels invoked.
The model's arithmetic intensity and throughput are calculated using the model's total flops and memory accesses.
This information is used to classify the entire model as either compute- or memory-bound.

Figure~\ref{fig:model_roofline} visualizes the roofline analysis for \texttt{MLPerf\_ResNet50\_v1.5} across batch sizes on \texttt{Tesla\_V100}.
We see that the model is compute-bound except for batch sizes $16$ and $32$ where it is memory-bound.
Looking into the data in
\analysis{2,8,10}
we find that the kernels invoked for the convolution layers sometimes vary across batch sizes.
This is because the cuDNN library relies on heuristics to choose the algorithm used for a convolution layer. 
The heuristics depend on the layer input parameters, available memory, etc.
For batch sizes less than $16$, 
the cuDNN convolution API uses the \texttt{IMPLICIT\_\allowbreak GEMM} algorithm and invokes the GPU kernel 
\texttt{cudnn::\allowbreak detail::\allowbreak implicit\_\allowbreak  convolve\_\allowbreak sgemm}.
This kernel has high arithmetic intensity and dominates the model's latency.
For batch sizes greater than $16$, the cuDNN convolution API chooses a different algorithm --- \texttt{IMPLICIT\_\allowbreak PRECOMP\_\allowbreak GEMM} algorithm, which invokes the GPU kernel \texttt{volta\_\allowbreak scudnn\_\allowbreak 128x64\_\allowbreak relu\_\allowbreak interior\_\allowbreak nn\_\allowbreak v1}.
Although this kernel is compute-bound, for batch sizes less than $64$ it has a relatively low arithmetic intensity.
Thus, for both batch sizes $16$ and $32$, this kernel's arithmetic intensity is not high enough to compensate for the effects of the other memory-bound kernels.
The result is that the overall model is memory-bound for batch sizes $16$ and $32$.
We also observe that the overall GPU achieved occupancy for the model increases as the batch size approaches the optimal batch size.

\analysis{8} and \analysis{10} are currently the most common types of analyses performed by researchers using NVIDIA's profilers.
Less common, but still possible, analyses without \carml{} are roofline analyses \analysis{9} and \analysis{15} as they require non-trivial scripts.
The scripts parse and aggregate the GPU profilers' outputs across multiple model evaluations to compute the roofline model.
Analyses \analysis{11-14} cannot be performed using existing tools as they require both the layer- and GPU kernel-level profiles and their results to be correlated.

%% file: sections/3.3-extensibility.tex
\subsection{Extensibility }\label{sec:extensibility}
Care was taken to ensure that \carml{}'s design is extensible.
Other profiling tools or methods can be integrated into \carml{} by implementing \carml{}'s tracer interface.
Thus, \carml{} can be extended with more tracers at each stack level or extended to capture more stack levels.
For example, one can integrate CPU profilers into \carml{} to capture both CPU and GPU information within the same timeline. 
One can also add a ML library profiling level between the layer- and GPU kernel-level to measure the cuDNN API calls.
Adding an application profiling level above the model level to measure whole applications (possibly distributed and using more than one ML model) is naturally supported by \carml{} as it uses distributed tracing.
As new profilers are introduced into \carml{}, one can add more types of analyses to the automated analysis pipeline.

%% file: sections/4.x_gpu_layer_info.tex
\begin{table}
    \centering
    \caption{The top $5$ most time-consuming layers in \analysis{11}~ for \texttt{MLPerf\_ResNet50\_v1.5} on \texttt{Tesla\_V100}.%
    }
    \resizebox{0.485\textwidth}{!}{%
        \begin{tabular}{lrcrrrrrrc} \toprule
        \centering%
         \shortstack{\textbf{Layer} \\ \textbf{Index}}  &
         \shortstack{\textbf{Layer} \\ \textbf{Latency} \\ \textbf{(ms)}} &
         \shortstack{\textbf{Kernel} \\ \textbf{Latency} \\ \textbf{(ms)}} & 
         \shortstack{\textbf{Layer} \\ \textbf{Gflops}}  &
         \shortstack{\textbf{Layer} \\ \textbf{DRAM} \\ \textbf{Reads} \\ \textbf{(MB)}} & \shortstack{\textbf{Layer} \\ \textbf{DRAM} \\ \textbf{Writes} \\ \textbf{(MB)}} & \shortstack{\textbf{Layer} \\ \textbf{Achieved} \\ \textbf{Occupancy} \\ \textbf{(\%)}}  &
         \shortstack{\textbf{Layer} \\ \textbf{Arithmetic} \\ \textbf{Intensity} \\ \textbf{(flops/byte)}}  &
         \shortstack{\textbf{Layer} \\ \textbf{Arithmetic} \\ \textbf{Throughput}\\ \textbf{(Tflops/s)}} &
         \shortstack{\textbf{Memory} \\ \textbf{Bound?}}
        \\   \midrule
208 & 7.59 & 7.45 & 79.74 & 362.67 & 548.50 & 19.43 & 83.46 & 10.70 & \no \\
221 & 7.57 & 7.43 & 79.74 & 368.11 & 551.70 & 19.43 & 82.68 & 10.73 & \no \\
195 & 5.67 & 5.55 & 59.20 & 36.51 & 17.99 & 15.80 & 1,036.10 & 10.67 & \no \\
3 & 5.08 & 4.91 & 62.89 & 11.55 & 284.21 & 13.23 & 202.78 & 12.80 & \no \\
113 & 4.67 & 4.57 & 59.22 & 76.65 & 21.36 & 15.31 & 576.17 & 12.94 & \no \\
        \bottomrule
        \end{tabular}%
	}
    \label{tab:gpu_layer_info}
\end{table}

%% file: sections/4.x_gpu_model_info.tex
\begin{table}
    \centering
    \caption{The \analysis{15} GPU kernel information aggregated within \texttt{MLPerf\_ResNet50\_v1.5} across batch sizes on \texttt{Tesla\_V100}.}
    \resizebox{0.48\textwidth}{!}{%
        \begin{tabular}{rrrrrrrc} \toprule
        \centering%
         \shortstack{\textbf{Batch} \\ \textbf{Size}}  &
         \shortstack{\textbf{Model} \\ \textbf{Latency} \\ \textbf{(ms)}} &
         \shortstack{\textbf{Kernel} \\ \textbf{Latency} \\ \textbf{(ms)}} & 
         \shortstack{\textbf{Model} \\ \textbf{Gflops}}  &
         \shortstack{\textbf{Model} \\ \textbf{DRAM} \\ \textbf{Reads} \\ \textbf{(MB)}} & \shortstack{\textbf{Model} \\ \textbf{DRAM} \\ \textbf{Writes} \\ \textbf{(MB)}} & \shortstack{\textbf{Model} \\ \textbf{Achieved} \\ \textbf{Occupancy} \\ \textbf{(\%)}}  &
         \shortstack{\textbf{Memory} \\ \textbf{Bound?}}
        \\   \midrule
1 & 6.21 & 5.01 & 7.94 & 192.49 & 194.16 & 22.65 & \no \\
2 & 6.83 & 5.93 & 16.08 & 290.41 & 354.54 & 22.47 & \no \\
4 & 8.51 & 7.68 & 30.95 & 659.11 & 720.15 & 26.39 & \no\\
8 & 12.80 & 11.60 & 60.66 & 1,676.07 & 1,496.81 & 31.97 & \no \\
16 & 21.90 & 20.14 & 118.04 & 3,969.19 & 3,024.09 & 35.58 & \yes \\
32 & 40.03 & 37.14 & 232.78 & 7,711.50 & 5,823.97 & 38.76 & \yes \\
64 & 74.03 & 67.72 & 429.08 & 10,932.22 & 9,268.27 & 43.18 & \no \\
128 & 142.89 & 131.79 & 873.63 & 16,071.32 & 16,105.40 & 44.48 & \no \\
256 & 275.05 & 254.25 & 1,742.39 & 23,185.11 & 31,095.45 & 43.15 &\no \\
        \bottomrule
        \end{tabular}%
	}
    \label{tab:gpu_model_info}
\end{table}

%% file: sections/5.0-systems_list.tex
\begin{table*}
    \caption{
    Five systems with Turing, Volta, Pascal, and Maxwell GPUs are selected for evaluation.
    We calculate the ideal arithmetic intensity of each system using the theoretic Flops and memory bandwidth reported by NVIDIA.
    }
    \centering
    \resizebox{0.9\textwidth}{!}{%
        \begin{tabular}{lcccrrr} \toprule
        \centering%
        \textbf{Name} & \textbf{CPU} & \textbf{GPU} & \shortstack{\textbf{GPU} \\ \textbf{Architecture}} & \shortstack{\textbf{Theoretical} \\ \textbf{FLOPS (TFLOPS)}} & \shortstack{\textbf{Memory Bandwidth} \\ \textbf{(GB/s)}}& \shortstack{\textbf{Ideal Arithmetic} \\ \textbf{Intensity (flops/byte)}} \\   \midrule
        Quadro\_RTX & Intel Xeon E5-2630 v4 @ 2.20GHz & Quadro RTX 6000 & Turing & 16.3 & 624 & 26.12 \\
        Tesla\_V100 (AWS P3) & Intel Xeon E5-2686 v4 @ 2.30GHz & Tesla V100-SXM2-16GB & Volta & 15.7 & 900 & 17.44 \\
        Tesla\_P100 & Intel Xeon E5-2682 v4 @ 2.50GHz & Tesla P100-PCIE-16GB & Pascal & 9.3 & 732 & 12.70 \\
        Tesla\_P4 & Intel Xeon E5-2682 v4 @ 2.50GHz & Tesla P4 & Pascal & 5.5 & 192 & 28.34 \\
        Tesla\_M60 (AWS G3) & Intel Xeon E5-2686 v4 @ 2.30GHz & Tesla M60 & Maxwell & 4.8 & 160 & 30.12 \\
        \bottomrule
        \end{tabular}%
	}%
    \label{tab:systems}
    \vspace{-20pt}
\end{table*}

%% file: sections/5.x-tf_models.tex
\begin{table*}
    \centering
    \caption{
    We use $55$ TensorFlow models from MLPerf, AI-Matrix, and TensorFlow Slim, Detection Zoo, DeepLab for evaluation.
    These models are sorted by the reported accuracy and solve different tasks: Image Classification (\textbf{IC}), Object Detection (\textbf{OD}), Instance Segmentation (\textbf{IS}), Semantic Segmentation (\textbf{SS}), and Super Resolution (\textbf{SR}).
    We measured the peak throughput achieved on \texttt{Tesla\_V100} and find the optimal batch size for each model.
    Online latency is defined as the model latency for batch size 1.
    Graph size is the size of the frozen graph for a model.
    }
    \resizebox{0.8\textwidth}{!}{%
\begin{tabular}{rlcrrrrrr} \toprule
\centering%
\textbf{ID} & \textbf{Name} & \textbf{Task} & \textbf{Accuracy}  &  \shortstack{\textbf{Graph Size} \\ \textbf{(MB)}} & \shortstack{ \textbf{Online} \\ \textbf{Latency (ms)}}  & \shortstack{ \textbf{Max Throughput} \\ \textbf{ (Inputs/Sec)}} & \shortstack{ \textbf{Optimal} \\ \textbf{Batch Size}} 
& \shortstack{ \textbf{Convolution} \\ \textbf{Percentage (\%)}} \\ \midrule
1 & Inception\_ResNet\_v2 & IC & 80.40 & 214 & 23.24 & 346.6 & 128 & 68.8 \\
2 & Inception\_v4 & IC & 80.20 & 163 & 17.29 & 436.7 & 128 & 75.7 \\
3 & Inception\_v3 & IC & 78.00 & 91 & 9.85 & 811.0 & 64 & 72.8 \\
4 & ResNet\_v2\_152 & IC & 77.80 & 231 & 14.05 & 466.8 & 256 & 60.5 \\
5 & ResNet\_v2\_101 & IC & 77.00 & 170 & 10.39 & 671.7 & 256 & 60.9 \\
6 & ResNet\_v1\_152 & IC & 76.80 & 230 & 13.70 & 541.3 & 256 & 69.6 \\
7 & MLPerf\_ResNet50\_v1.5 & IC & 76.46 & 103 & 6.22 & 930.7 & 256 & 58.7 \\
8 & ResNet\_v1\_101 & IC & 76.40 & 170 & 10.01 & 774.7 & 256 & 69.9 \\
9 & AI\_Matrix\_ResNet152 & IC & 75.93 & 230 & 14.61 & 468.0 & 256 & 61.8 \\
10 & ResNet\_v2\_50 & IC & 75.60 & 98 & 6.23 & 1,119.7 & 256 & 58.1 \\
11 & ResNet\_v1\_50 & IC & 75.20 & 98 & 6.19 & 1,284.6 & 256 & 67.5 \\
12 & AI\_Matrix\_ResNet50 & IC & 74.38 & 98 & 5.99 & 1,060.3 & 256 & 57.9 \\
13 & Inception\_v2 & IC & 73.90 & 43 & 6.45 & 2,032.0 & 128 & 68.2 \\
14 & AI\_Matrix\_DenseNet121 & IC & 73.29 & 31 & 12.80 & 846.4 & 32 & 49.3 \\
15 & MLPerf\_MobileNet\_v1 & IC & 71.68 & 17 & 3.15 & 2,576.4 & 128 & 52.0 \\
16 & VGG16 & IC & 71.50 & 528 & 21.33 & 687.5 & 256 & 74.7 \\
17 & VGG19 & IC & 71.10 & 548 & 22.10 & 593.4 & 256 & 76.7 \\
18 & MobileNet\_v1\_1.0\_224 & IC & 70.90 & 16 & 3.19 & 2,580.6 & 128 & 51.9 \\
19 & AI\_Matrix\_GoogleNet & IC & 70.01 & 27 & 5.35 & 2,464.5 & 128 & 62.9 \\
20 & MobileNet\_v1\_1.0\_192 & IC & 70.00 & 16 & 3.11 & 3,460.8 & 128 & 52.5 \\
21 & Inception\_v1 & IC & 69.80 & 26 & 5.30 & 2,576.6 & 128 & 63.7 \\
22 & BVLC\_GoogLeNet\_Caffe & IC & 68.70 & 27 & 6.53 & 951.7 & 8 & 55.1 \\
23 & MobileNet\_v1\_0.75\_224 & IC & 68.40 & 10 & 3.18 & 3,183.7 & 64 & 51.1 \\
24 & MobileNet\_v1\_1.0\_160 & IC & 68.00 & 16 & 3.01 & 4,240.5 & 64 & 55.4 \\
25 & MobileNet\_v1\_0.75\_192 & IC & 67.20 & 10 & 3.05 & 4,187.8 & 64 & 51.8 \\
26 & MobileNet\_v1\_0.75\_160 & IC & 65.30 & 10 & 2.81 & 5,569.6 & 64 & 53.1 \\
27 & MobileNet\_v1\_1.0\_128 & IC & 65.20 & 16 & 2.91 & 6,743.2 & 64 & 55.9 \\
28 & MobileNet\_v1\_0.5\_224 & IC & 63.30 & 5.2 & 3.55 & 3,346.5 & 64 & 63.0 \\
29 & MobileNet\_v1\_0.75\_128 & IC & 62.10 & 10 & 2.96 & 8,378.4 & 64 & 55.7 \\
30 & MobileNet\_v1\_0.5\_192 & IC & 61.70 & 5.2 & 3.28 & 4,453.2 & 64 & 63.3 \\
31 & MobileNet\_v1\_0.5\_160 & IC & 59.10 & 5.2 & 3.22 & 6,148.7 & 64 & 63.7 \\
32 & BVLC\_AlexNet\_Caffe & IC & 57.10 & 233 & 2.33 & 2,495.8 & 16 & 36.3 \\
33 & MobileNet\_v1\_0.5\_128 & IC & 56.30 & 5.2 & 3.20 & 8,924.0 & 64 & 64.1 \\
34 & MobileNet\_v1\_0.25\_224 & IC & 49.80 & 1.9 & 3.40 & 5,257.9 & 64 & 60.6 \\
35 & MobileNet\_v1\_0.25\_192 & IC & 47.70 & 1.9 & 3.26 & 7,135.7 & 64 & 61.2 \\
36 & MobileNet\_v1\_0.25\_160 & IC & 45.50 & 1.9 & 3.15 & 10,081.5 & 256 & 68.4 \\
37 & MobileNet\_v1\_0.25\_128 & IC & 41.50 & 1.9 & 3.15 & 10,707.6 & 256 & 80.2 \\ \hdashline
38 & Faster\_RCNN\_NAS & OD & 43 & 405 & 5079.32 & 0.6 & 4 & 85.2 \\
39 & Faster\_RCNN\_ResNet101 & OD & 32 & 187 & 91.15 & 14.67 & 4 & 13 \\
40 & SSD\_MobileNet\_v1\_FPN & OD & 32 & 49 & 47.44 & 33.46 & 8 & 4.8 \\
41 & Faster\_RCNN\_ResNet50 & OD & 30 & 115 & 81.19 & 16.49 & 4 & 10.8 \\
42 & Faster\_RCNN\_Inception\_v2 & OD & 28 & 54 & 61.88 & 22.17 & 4 & 4.7 \\
43 & SSD\_Inception\_v2 & OD & 24 & 97 & 50.34 & 32.26 & 8 & 2.5 \\
44 & MLPerf\_SSD\_MobileNet\_v1\_300x300 & OD & 23 & 28 & 47.49 & 33.51 & 8 & 0.8 \\
45 & SSD\_MobileNet\_v2 & OD & 22 & 66 & 48.72 & 32.4 & 8 & 1.3 \\
46 & MLPerf\_SSD\_ResNet34\_1200x1200 & OD & 20 & 81 & 87.4 & 11.44 & 1 & 14.9 \\
47 & SSD\_MobileNet\_v1\_PPN & OD & 20 & 10 & 47.07 & 33.1 & 16 & 0.6 \\ \hdashline
48 & Mask\_RCNN\_Inception\_ResNet\_v2 & IS & 36 & 254 & 382.52 & 2.92 & 4 & 29.2 \\
49 & Mask\_RCNN\_ResNet101\_v2 & IS & 33 & 212 & 295.18 & 3.6 & 2 & 42.4 \\
50 & Mask\_RCNN\_ResNet50\_v2 & IS & 29 & 138 & 231.22 & 4.64 & 2 & 40.3 \\
51 & Mask\_RCNN\_Inception\_v2 & IS & 25 & 64 & 86.86 & 17.25 & 4 & 5.7 \\ \hdashline
52 & DeepLabv3\_Xception\_65 & SS & 87.8 & 439 & 72.55 & 13.78 & 1 & 49.2 \\
53 & DeepLabv3\_MobileNet\_v2 & SS & 80.25 & 8.8 & 10.96 & 91.27 & 1 & 42.1 \\
54 & DeepLabv3\_MobileNet\_v2\_DM0.5 & SS & 71.83 & 7.6 & 9.5 & 105.21 & 1 & 41.5 \\ \hdashline
55 & SRGAN & SR & - & 5.9 & 70.29 & 14.23 & 1 & 62.3 \\
\bottomrule
\end{tabular}%
	}
    \label{tab:tf_models}
    \vspace{-20pt}
\end{table*}

%% file: sections/5.xx-tf_models_cls.tex
\begin{table*}
    \centering
    \caption{In-depth characterization of the $37$ image classification models listed in Table~\ref{tab:tf_models} at the optimal batch sizes on \texttt{Tesla\_v100}. The model execution is partitioned into beginning (\textit{B}), middle (\textit{M}) , and end (\textit{E}) intervals based on layer index. 
    The most intensive stages for latency, memory allocation, flops and memory access are shown.
    }
    \resizebox{0.85\textwidth}{!}{%
\begin{tabular}{rrrrrrrrrrccccc}
\hline
\textbf{ID} & \shortstack{ \textbf{Batch} \\ \textbf{Latency} \\ \textbf{(ms)}} & \shortstack{ \textbf{GPU} \\ \textbf{Latency} \\ \textbf{Percentage} \\ \textbf{(\%)}} & \shortstack{\textbf{GPU} \\ \textbf{Gflops}} & \shortstack{ \textbf{GPU} \\ \textbf{DRAM} \\ \textbf{Read} \\ \textbf{(GB)}} & \shortstack{\textbf{GPU} \\ \textbf{DRAM} \\ \textbf{Write} \\ \textbf{(GB)}} & \shortstack{ \textbf{GPU} \\ \textbf{ Achieved} \\ \textbf{Occupancy} \\ \textbf{(\%)}} & \shortstack{\textbf{Arithmetic} \\ \textbf{Intensity}  \\ \textbf{(Flops/byte)}} & \shortstack{ \textbf{Arithmetic} \\ \textbf{Throughput} \\ \textbf{(TFlops)}} & \shortstack{ \textbf{Memory} \\ \textbf{Bound?}}
& \shortstack{ \textbf{Latency} \\ \textbf{Stage}}
&  \shortstack{ \textbf{Allocated} \\
 \textbf{Memory} \\ \textbf{Stage}} 
&  \shortstack{ \textbf{flops} \\ \textbf{Stage}} 
&  \shortstack{ \textbf{Memory} \\
 \textbf{Access} \\ \textbf{Stage}} \\
\hline
1 & 400.06 & 94.77 & 2,910.44 & 50.64 & 38.74 & 39.74 & 32.56 & 7.68 & \no & M & M & M & M \\
2 & 324.49 & 93.92 & 2,492.92 & 27.25 & 24.48 & 33.79 & 48.19 & 8.18 & \no & M & M & M & M \\
3 & 86.39 & 88.05 & 552.22 & 10.54 & 8.18 & 34.6 & 29.50 & 7.26 & \no & M & M & M & B \\
4 & 593.97 & 96.32 & 3,954.06 & 58.90 & 65.44 & 43.51 & 31.80 & 6.91 & \no & E & E & M & E \\
5 & 412.37 & 94.90 & 2,725.14 & 39.08 & 44.62 & 42.88 & 32.56 & 6.96 & \no & E & E & M & E \\
6 & 517.11 & 95.90 & 3,947.38 & 51.17 & 54.77 & 42.78 & 37.26 & 7.96 & \no & E & E & M & E \\
7 & 275.05 & 92.43 & 1,742.39 & 24.40 & 32.61 & 43.15 & 30.62 & 6.85 & \no & B & E & M & E \\
8 & 360.90 & 94.29 & 2,720.62 & 33.87 & 37.12 & 42.19 & 38.32 & 7.99 & \no & E & E & M & E \\
9 & 591.47 & 96.29 & 4,034.74 & 63.70 & 72.16 & 43.9 & 29.70 & 7.08 & \no & B & M & B & M \\
10 & 245.07 & 91.74 & 1,480.10 & 21.84 & 28.29 & 42.96 & 29.52 & 6.58 & \no & E & E & M & E \\
11 & 213.52 & 90.42 & 1,477.33 & 18.79 & 22.76 & 42.29 & 35.56 & 7.65 & \no & E & E & M & E \\
12 & 257.80 & 91.89 & 1,561.76 & 24.86 & 33.39 & 44.26 & 26.81 & 6.59 & \no & B & M & B & M \\
13 & 68.27 & 83.62 & 363.33 & 9.67 & 7.32 & 40.23 & 21.38 & 6.36 & \no & B & B & M & B \\
14 & 40.24 & 93.32 & 150.02 & 10.13 & 7.93 & 44.94 & 8.30 & 4.00 & \yes & B & B & B & B \\
15 & 51.57 & 79.76 & 148.18 & 7.08 & 6.81 & 52.58 & 10.67 & 3.60 & \yes & M & M & M & M \\
16 & 399.31 & 94.98 & 2,655.39 & 24.38 & 33.23 & 26.14 & 46.10 & 7.00 & \no & B & B & M & E \\
17 & 464.47 & 95.61 & 3,207.02 & 26.44 & 37.65 & 24.91 & 50.04 & 7.22 & \no & B & B & M & E \\
18 & 51.59 & 79.73 & 148.18 & 6.97 & 6.75 & 52.59 & 10.80 & 3.60 & \yes & M & M & M & M \\
19 & 56.08 & 80.20 & 259.14 & 7.63 & 6.18 & 42.16 & 18.76 & 5.76 & \no & M & B & M & B \\
20 & 38.48 & 79.55 & 108.93 & 6.51 & 6.19 & 52.32 & 8.58 & 3.56 & \yes & M & M & M & B \\
21 & 53.35 & 79.43 & 252.06 & 7.21 & 5.61 & 41.74 & 19.67 & 5.95 & \no & M & B & M & B \\
22 & 9.08 & 80.00 & 20.26 & 0.73 & 0.84 & 33.87 & 12.97 & 2.79 & \yes & E & B & E & B \\
23 & 20.82 & 73.14 & 45.10 & 4.86 & 4.11 & 52.73 & 5.03 & 2.96 & \yes & M & M & M & M \\
24 & 14.92 & 78.26 & 38.17 & 3.24 & 2.88 & 48.92 & 6.23 & 3.27 & \yes & M & M & M & M \\
25 & 15.69 & 72.61 & 33.10 & 3.52 & 3.08 & 52.02 & 5.01 & 2.91 & \yes & M & M & M & M \\
26 & 11.30 & 71.86 & 23.14 & 2.31 & 2.17 & 51.01 & 5.17 & 2.85 & \yes & M & M & M & M \\
27 & 9.86 & 77.23 & 24.39 & 1.90 & 1.84 & 47.78 & 6.54 & 3.20 & \yes & M & M & M & M \\
28 & 20.00 & 71.93 & 52.03 & 2.99 & 2.85 & 43.87 & 8.91 & 3.62 & \yes & B & M & B & M \\
29 & 7.75 & 71.35 & 14.80 & 1.26 & 1.35 & 47.12 & 5.68 & 2.68 & \yes & M & M & M & M \\
30 & 15.07 & 71.75 & 38.22 & 2.08 & 2.09 & 43.27 & 9.17 & 3.53 & \yes & B & M & B & M \\
31 & 10.91 & 71.38 & 26.62 & 1.29 & 1.42 & 41.43 & 9.83 & 3.42 & \yes & B & M & B & M \\
32 & 6.52 & 68.69 & 15.36 & 0.76 & 0.51 & 37.31 & 12.11 & 3.43 & \yes & B & B & B & B \\
33 & 7.44 & 70.48 & 17.05 & 0.71 & 0.88 & 39.88 & 10.73 & 3.25 & \yes & B & M & B & M \\
34 & 11.95 & 53.93 & 14.79 & 1.25 & 1.42 & 44.25 & 5.52 & 2.30 & \yes & B & M & B & M \\
35 & 9.09 & 53.68 & 10.87 & 0.84 & 1.02 & 43.46 & 5.82 & 2.23 & \yes & B & M & B & M \\
36 & 25.36 & 60.78 & 36.75 & 3.26 & 3.09 & 42.39 & 5.79 & 2.38 & \yes & B & M & B & M \\
37 & 23.71 & 70.01 & 23.81 & 1.87 & 2.31 & 39.8 & 5.69 & 1.43 & \yes & M & M & B & M \\
\hline
\end{tabular}%
	}
    \label{tab:tf_models_cls}
    \vspace{-10pt}
\end{table*}

%% file: sections/5-evaluation.tex
\begin{figure}
  \centering
  \includegraphics[width=0.48\textwidth]{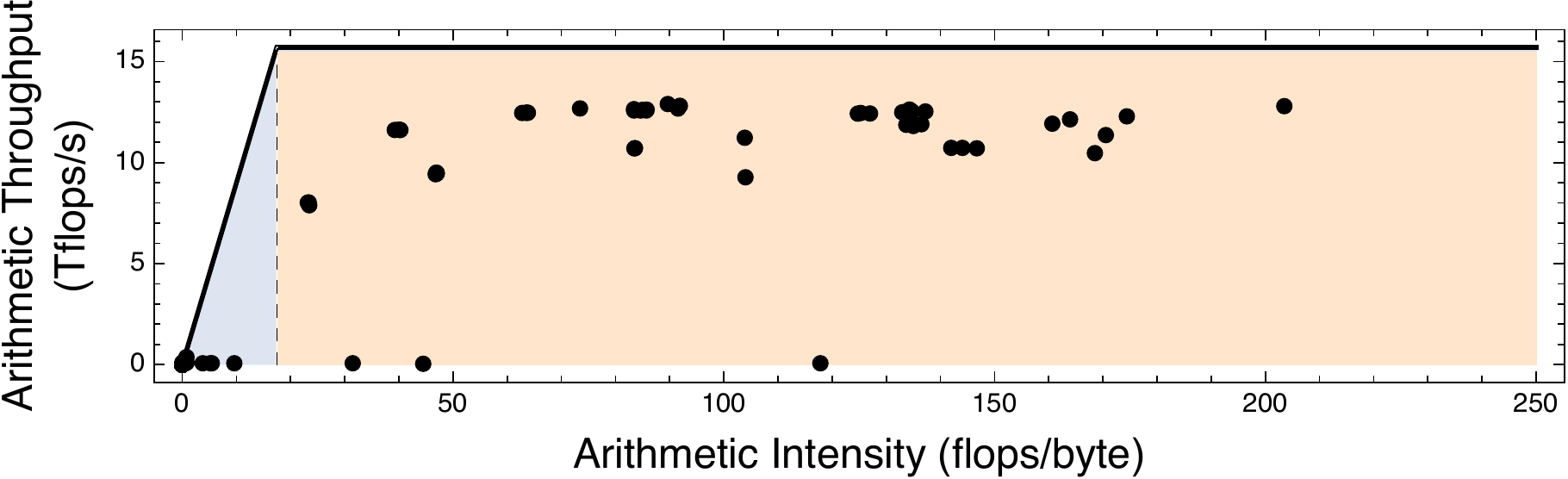}
  \caption{The \analysis{14}~ roofline analysis for all the layers in \texttt{MLPerf\_ResNet50\_v1.5} with batch size $256$ on \texttt{Tesla\_V100}.%
  }
  \label{fig:layer_roofline}
\end{figure}

\begin{figure}
  \centering
  \includegraphics[width=0.48\textwidth]{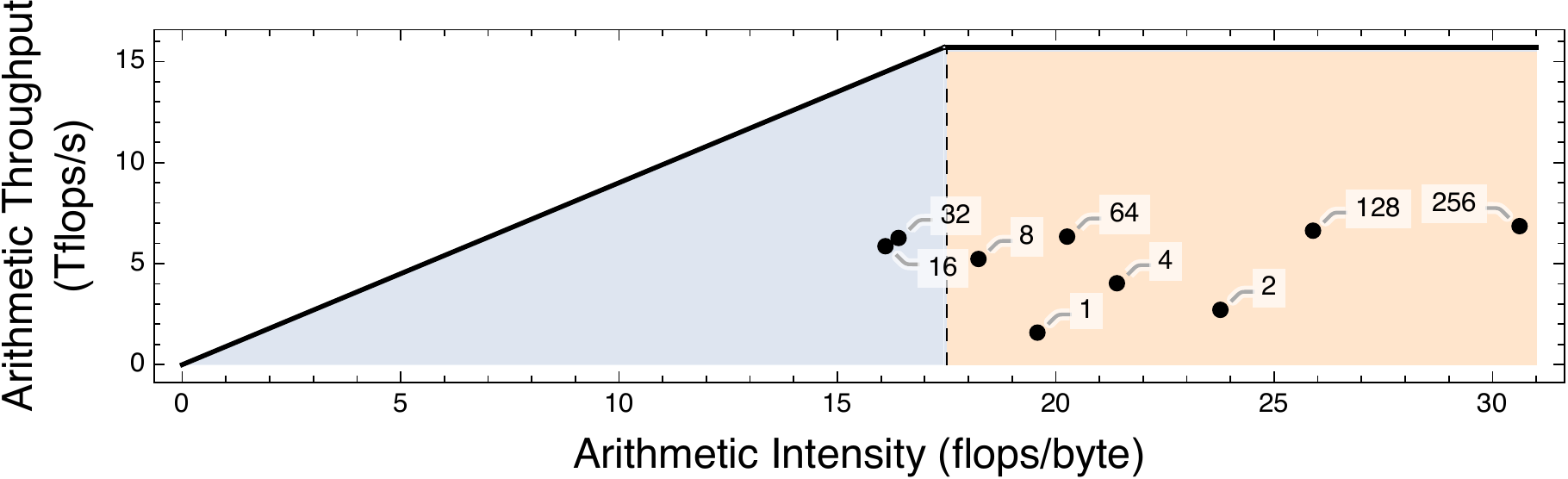}
  \caption{The roofline analysis for \texttt{MLPerf\_ResNet50\_v1.5} across batch sizes on \texttt{Tesla\_V100} using \analysis{15}~.%
  }
  \label{fig:model_roofline}
\end{figure}

\section{Evaluation}\label{sec:evaluation}

We profile and characterize $55$ state-of-the-art TensorFlow ML models (Table~\ref{tab:tf_models}) selected from the MLPerf Inference~\cite{mlperf}, AI-Matrix~\cite{aimatrix}, and TensorFlow model zoo~\cite{tfslimzoo, tfodzoo, tfdlzoo}.
The models solve computer vision tasks including image classification, object detection, instance segmentation, semantic segmentation, and super resolution.
To compare TensorFlow against MXNet, we select an additional $10$ MXNet models from the MXNet Gluon model zoo~\cite{mxzoo} (Table~\ref{tab:mx_models_cls}) that are comparable to the TensorFlow models.
We evaluated the models using NGC TensorFlow container v$19.06$, and NGC MXNet container v$19.06$ on $5$ representative GPU systems listed in Table~\ref{tab:systems}.
This section presents insights about the models, frameworks, and GPU systems using the \carml{}'s analyses described in Section~\ref{sec:analysis}.

%% file: sections/5.xxx-mx_models_cls.tex
\begin{table*}[ht]
    \centering
    \caption{Characterization of $10$ MXNet models, which are comparable to the TensorFlow ones listed in Table~\ref{tab:tf_models} (labeled with the same ID). The online latency is measured at batch size 1 and the others are measured at the model's optimal batch size on \texttt{Tesla\_V100}. The online latency and maximum throughput are normalized to TensorFlow's.}
    \label{tab:mx_models_cls}
    \resizebox{0.9\textwidth}{!}{%
\begin{tabular}{rlrrrrrrrrrrc}
\hline
\textbf{ID} & \textbf{Name} & \shortstack{ \textbf{Normalized} \\ \textbf{Online} \\ \textbf{Latency}} & \shortstack{ \textbf{Optimal} \\ \textbf{Batch} \\ \textbf{Size}} & \shortstack{ \textbf{Normalized} \\ \textbf{Maximum} \\ \textbf{Throughput}} & \shortstack{ \textbf{GPU} \\ \textbf{Latency} \\ \textbf{Percentage}}  & \shortstack{\textbf{GPU} \\ \textbf{Gflops}} & \shortstack{ \textbf{GPU} \\ \textbf{DRAM} \\ \textbf{Read} \\ \textbf{(GB)}} & \shortstack{\textbf{GPU} \\ \textbf{DRAM} \\ \textbf{Write} \\ \textbf{(GB)}} & \shortstack{ \textbf{GPU} \\ \textbf{ Achieved} \\ \textbf{Occupancy} \\ \textbf{(\%)}} & \shortstack{\textbf{Arithmetic} \\ \textbf{Intensity}  \\ \textbf{(Flops/byte)}} & \shortstack{ \textbf{Arithmetic} \\ \textbf{Throughput} \\ \textbf{(TFlops)}} & \shortstack{ \textbf{Memory} \\ \textbf{Bound?}} \\
\hline
4 & ResNet\_v2\_152 & 1.76 & 256 & 1.03 & 97.00 & 4,116.42 & 49.05 & 52.62 & 46.91 & 38.61 & 7.95 & \no \\
5 & ResNet\_v2\_101 & 1.59 & 256 & 1.02 & 96.77 & 2,882.65 & 32.33 & 36.16 & 46.38 & 40.14 & 7.96 & \no \\
6 & ResNet\_v1\_152 & 1.68 & 256 & 0.90 & 96.20 & 3,828.11 & 51.29 & 55.00 & 49.40 & 34.35 & 7.54 & \no \\
8 & ResNet\_v1\_101 & 1.60 & 256 & 0.91 & 95.67 & 2,589.76 & 33.93 & 37.84 & 49.57 & 34.42 & 7.45 & \no \\
10 & ResNet\_v2\_50 & 1.41 & 256 & 1.03 & 97.10 & 1,636.10 & 17.03 & 22.60 & 46.98 & 39.37 & 7.60 & \no \\
11 & ResNet\_v1\_50 & 1.32 & 256 & 0.96 & 94.90 & 1,339.50 & 18.37 & 24.04 & 51.97 & 30.12 & 6.76 & \no \\
18 & MobileNet\_v1\_1.0\_224 & 1.00 & 256 & 1.54 & 93.75 & 298.38 & 6.91 & 8.29 & 63.53 & 18.71 & 4.96 & \no \\
23 & MobileNet\_v1\_0.75\_224 & 0.95 & 64 & 1.76 & 79.49 & 45.00 & 3.47 & 2.73 & 63.38 & 6.92 & 4.08 & \yes \\
28 & MobileNet\_v1\_0.5\_224 & 0.87 & 64 & 1.35 & 81.01 & 51.47 & 1.99 & 1.82 & 48.68 & 12.88 & 4.49 & \yes \\
34 & MobileNet\_v1\_0.25\_224 & 0.93 & 64 & 1.64 & 64.32 & 13.77 & 0.81 & 0.90 & 50.57 & 7.64 & 2.88 & \yes \\
\hline
\end{tabular}%
	}
\end{table*}

%% file: sections/5.1-model.tex
\subsection{Model Evaluation}\label{sec:eval_model}

Using the model- and layer-level profiling data, we look at all $55$ TensorFlow models in Table~\ref{tab:tf_models}.
Models solving the same task are clustered together and are then sorted by their reported accuracy.
The table shows each model's accuracy, model graph size, online latency (batch size is $1$), maximum throughput, optimal batch size (described in Section~\ref{sec:modellvl}), and percentage of latency  attributed to convolution layers.

\textbf{Model latency percentage of convolution layers} ---
Using the model- and layer-level profile data, we calculate the percentage of model latency attributed to convolution layers (Tensorflow's \texttt{Conv2D} and \texttt{DepthwiseConv2dNative} layers) with each model's optimal batch size on \texttt{Tesla\_V100}.
This is shown in the last column of Table~\ref{tab:tf_models}.
We observe that:
\circledwhite{1} the convolution layer latency percentage ranges between $36.3\%$ and $80.2\%$ for image classification models.
This suggests that convolution layers still dominate (but not exclusively) the latency of image classification models --- even on recent GPUs.
This is not true for \circledwhite{2} object detection models, which (except for \texttt{Faster\_RCNN\_NAS}) attribute only $0.6\%$ to $14.9\%$ of latency to convolution layers.
For these models, the dominating layer type is \texttt{Where}, which reshapes a tensor with respect to a user-defined operator.
For \circledwhite{3} instance segmentation models, convolution layers dominate the model latency; except for \texttt{Mask\_RCNN\_Inception\_v2} whose latency is also dominated by \texttt{Where} layers.
For \circledwhite{4} semantic segmentation models, the model latency is affected by both the convolution layers  and the memory-bound layers (such as \texttt{Transpose}, \texttt{Add}, and \texttt{Mul}).
Finally, \circledwhite{5} the super resolution model \texttt{SRGAN} is dominated by convolution layers.

\begin{figure}
  \centering
    \setlength{\abovecaptionskip}{0pt}
  \includegraphics[width=0.48\textwidth]{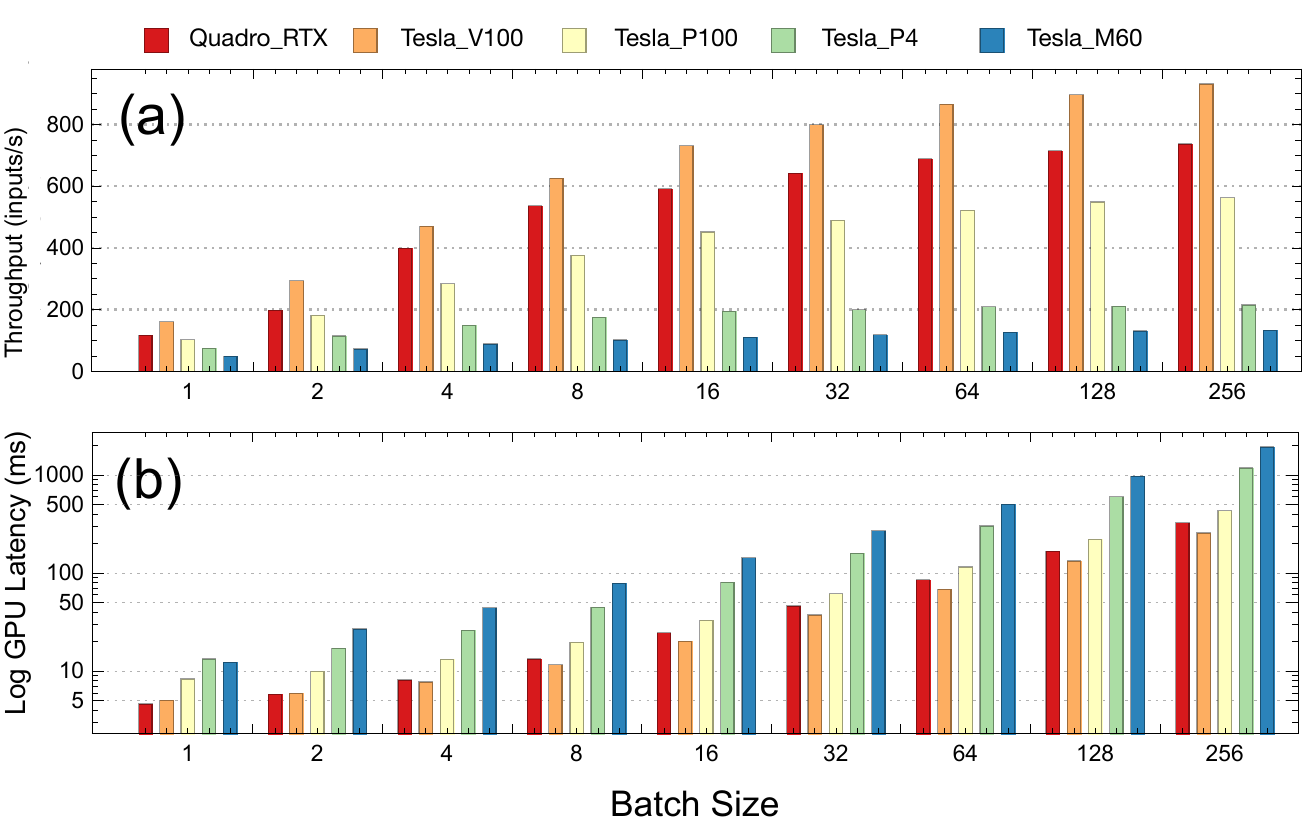}
  \caption{The throughput and latency (log scale) of \texttt{MLPerf\_ResNet50\_v1.5} across batch sizes and systems.
  }
  \label{fig:machines_throughput_latency}
\end{figure}

\textbf{GPU latency, flops and memory accesses} ---
Using the model-, layer-, and GPU kernel-level profiling, we perform an in-depth analyses of the $37$ image classification models at their optimal batch sizes on \texttt{Tesla\_V100}.
Table~\ref{tab:tf_models_cls} shows the model’s latency at the optimal batch size, GPU latency percentage (i.e. the latency due to GPU kernel execution normalized to the model latency), GPU metrics, and arithmetic intensity and throughput.
It also shows the most intensive stage for latency, memory allocation, GPU flops, and memory access throughout the model execution.
We find that across the models the GPU latency percentage varies from $53.68\%$ to $95.61\%$ and is roughly proportional to the number of flops and memory accesses (the sum of GPU DRAM reads and writes).
We also observe that models with high batch latency tend to have a high GPU latency percentage.
This either suggests that the GPU saturates for these models or that the models are not well optimized for GPU execution.
The low GPU latency percentage for some models shows that the time spent within non-GPU code (framework overhead, GPU stalls due to synchronization, etc.) is high.

\textbf{Batch size vs GPU achieved occupancy} ---
The GPU achieved occupancy is a partial indicator of GPU utilization.
Table~\ref{tab:gpu_model_info} shows that as a model's batch size approaches the optimal, its overall achieved GPU occupancy increases.

\begin{figure}
  \centering
    \setlength{\abovecaptionskip}{0pt}
  \includegraphics[width=0.48\textwidth]{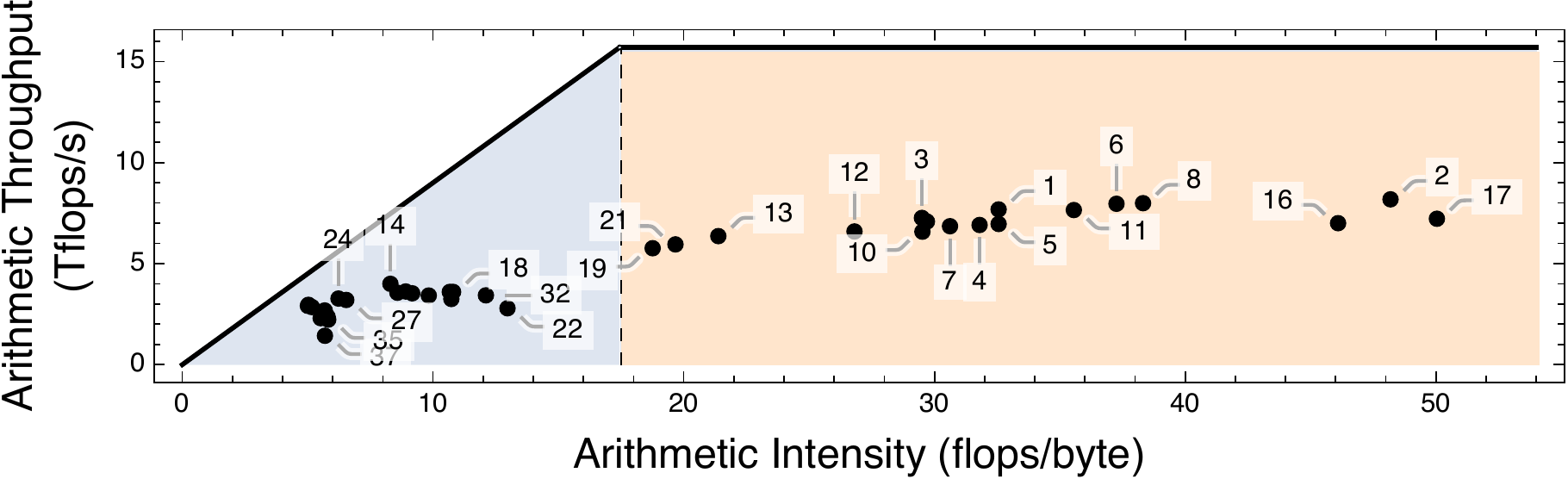}
  \caption{The roofline analysis for the $37$ image classification models with their the optimal batch sizes on \texttt{Tesla\_V100}.
  }
  \label{fig:models_roofline}
\end{figure}

\textbf{Roofline analysis} ---
Figure~\ref{fig:models_roofline} shows the roofline analysis for all $37$ image classification models with their optimal batch sizes on \texttt{Tesla\_V100}.
Out of $37$ models, $20$ are memory-bound.
Models with low compute and memory requirements tend to be memory-bound and have lower accuracy, e.g. some variants of \texttt{MobileNet} which target edge devices. 
All models achieve at most $52\%$ of the theoretical peak throughput, suggesting that there is room for  optimizations.

\textbf{Latency, memory allocation, flops, and memory access trend} --- 
To understand the performance trend within model execution, we divide the model execution into $3$ intervals based on the layer index: beginning, middle, and end based on the layer index.
We then compute the total latency, flops, and memory accesses within each interval and identify which interval dominates.
The last $4$ columns in Table~\ref{tab:tf_models_cls} show the results of the $37$ image classification models on \texttt{Tesla\_V100}.
The demanding intervals vary across models and suggest that one can potentially interleave multiple model executions to increase GPU utilization.

%% file: sections/5.2-framework.tex
\subsection{ML Framework Evaluation}\label{sec:eval_framework}

To compare ML frameworks, $10$ MXNet models are selected from the MXNet model zoo~\cite{mxzoo}.
We choose $6$ variants of \texttt{ResNet} which are compute-intensive and are compute-bound (at the optimal batch size), and $4$ variants \texttt{MobileNet} which are less compute-intensive and are memory-bound.
The models (shown in Table~\ref{tab:mx_models_cls}) are comparable to the TensorFlow models.
We perform the comparison between the TensorFlow and MXNet frameworks on \texttt{Tesla\_V100}.
The online latency and maximum throughput in the Table~\ref{tab:mx_models_cls} are normalized to the corresponding values using TensorFlow.
We use \carml{} to compute the optimal batch size for each MXNet model.
Except for model $18$, the optimal batch size for all MXNet models match the corresponding TensorFlow models.

\textbf{Compute-bound models} ---
Table~\ref{tab:mx_models_cls} shows that the online latency (batch size $1$) of MXNet \texttt{ResNets} is higher than that of the corresponding TensorFlow model.
After looking into the analysis results, we find that while the total GPU kernel latencies of TensorFlow and MXNet \texttt{ResNets} are about the same, the MXNet \texttt{ResNets} have a much higher non-GPU latency.
MXNet \texttt{ResNet\_v1\_50}, for example, has a non-GPU latency of $4.44ms$ ($55.1\%$ of the total online latency) whereas it is only $2.18ms$ for TensorFlow \texttt{ResNet\_v1\_50} ($35.3\%$ of the total).
We observe that as the batch size increases (and the model becomes more compute-bound) the percentage of the non-GPU latency decreases and MXNet \texttt{ResNets} achieve about the same maximum throughput as TensorFlow \texttt{ResNets}.
At the optimal batch size, TensorFlow and MXNet \texttt{ResNets} have comparable GPU latency percentage, flops, memory accesses, achieved occupancy, and roofline results.
This suggests that MXNet incurs a fixed overhead for model execution which is more pronounced for small batch sizes.

\textbf{Memory-bound models} ---
For the less compute-intensive \texttt{MobileNets}, we observe that MXNet the \texttt{MobileNets} achieve the same online latency as the corresponding TensorFlow model.
However, as the batch size increases (and the models become memory-bound) we find that MXNet \texttt{MobileNets} has fewer memory accesses and therefore a higher achieved GPU occupancy compared to the TensorFlow models.
As a result, MXNet\texttt{MobileNets} achieve between $35\%$ and $74\%$ more throughput at their optimal batch sizes (shown in Table~\ref{tab:mx_models_cls}).
Further GPU kernel-level analysis attributes the cause to the Eigen library.
The Eigen library is used by TensorFlow (but not MXNet) for element-wise layers and it incurs excessive DRAM reads and writes.
This becomes a performance-limiting factor for memory-bound models.

%% file: sections/5.3-system.tex
\subsection{System Evaluation}\label{sec:eval_gpu}

We use \carml{} to evaluate \texttt{MLPerf\_\allowbreak ResNet50\_\allowbreak  v1.5} on all $5$ GPU systems in Table~\ref{tab:systems} using the NGC TensorFlow container. 
We fix the software stack (TensorFlow, cuDNN, cuBLAS, CUDA version, etc.) on all $5$ systems to be the same.
Figure~\ref{fig:machines_throughput_latency}a shows the throughput across systems and batch sizes.
Figure~\ref{fig:machines_throughput_latency}b shows the GPU latency (the total latency of all the GPU kernel calls) in log scale for the $5$ systems across batch sizes.
Although the \texttt{Quadro\_RTX} GPU has a slightly higher peak FLOPS compared to \texttt{Tesla\_V100}, it has a much lower memory bandwidth.
Hence, \texttt{Quadro\_RTX} straggles on memory-bound layers and performs slightly worse when compared to \texttt{Tesla\_V100}.
We observe that the performance at each batch size differs across systems.
The performance also scales differently across systems with respect to the batch size.

Looking at the GPU kernel-level profile for each system, we find that 
the GPU kernels invoked are system-dependent --- even with the same batch size and software stack.
Both \texttt{Quadro\_RTX} and \texttt{Tesla\_V100} call the same set of GPU kernels, while the other $3$ systems use a different set of GPU kernels.
This is because the same cuDNN API may use different GPU kernels for different GPU systems.
For example, the convolution layers for batch size $256$ on \texttt{Tesla\_P100}, \texttt{Tesla\_P4}, and \texttt{Tesla\_M60} invoke the \texttt{maxwell\_scudnn\_*} kernels, whereas on \texttt{Quadro\_RTX} and \texttt{Tesla\_V100} the \texttt{volta\_scudnn\_*} kernels are invoked.
This implies that cuDNN uses optimized kernels for GPU generations after Volta.
Furthermore, because of the cuDNN algorithm selection heuristics, the distribution of the kernel calls differs across systems.
For example, \texttt{Tesla\_V100} calls the \texttt{volta\_scudnn\_128x64\_relu\_interior\_nn\_v1} kernel $34$ times whereas \texttt{Quadro\_RTX} calls it $18$ times  (the other $16$ being dispatched to the \texttt{volta\_scudnn\_128x128\_relu\_interior\_nn\_v1} kernel).

%% file: sections/7-conclusion.tex
\section{Conclusion}\label{sec:conclusion}

A big hurdle in optimizing and deploying ML workloads is understanding their performance characteristics across the HW/SW stack.
The analyses currently performed on ML models and systems are largely limited by the lack of correlation between profiles from different profiling tools or methods.
This paper proposes \carml, an across-stack profiling design that aggregates profile data from different sources and correlates them to construct a holistic and hierarchical view of ML model execution.
While the across-stack profiling design is general, this paper focuses on how it enables in-depth automated profiling and characterization of ML models on GPUs.
We use \carml{}'s profiling and analysis capabilities to systematically characterize $65$ state-of-the-art ML models.
Through the $15$ types of analysis introduced, we derive meaningful insights that would otherwise be difficult to discern without \carml{}.
We show that \carml helps researchers understand the sources of inefficiency in ML models, frameworks, and systems.

%% file: sections/99-ack.tex
\vspace{10pt}

\section*{Acknowledgments}
\label{sec:ack}

This work is supported by the IBM-ILLINOIS Center for Cognitive Computing Systems Research (C3SR) - a member of the IBM Cognitive Horizon Network, and the Applications Driving Architectures (ADA) Research Center - one of the JUMP Centers co-sponsored by SRC and DARPA.